\documentclass[preprint,12pt]{elsarticle}

\usepackage[
    left=1.7cm,
    right=1.7cm,
    top=1.8cm,
    bottom=1.8cm
]{geometry}
\usepackage{amssymb}
\usepackage{longtable}
\usepackage{tabularx}
\usepackage{array}
\usepackage{makecell}
\usepackage{xurl}
\usepackage{xurl}
\usepackage{float}
\usepackage{booktabs}

\usepackage{amsmath}

\begin{document}

\begin{frontmatter}

\title{Fairness or Fallacy? Mapping the Landscape of AI Bias in Disease Detection}

\author[label1]{Ghulam Mudassir}
\ead{ghulam.mudassir@buckingham.ac.uk}

\author[label2]{Hameed Ullah Khan}
\ead{hameed.khan@case.edu.pk}

\author[label3]{Ahmad Sami Al-Shamayleh}

\ead{a.alshamayleh@ammanu.edu.jo}
\author[label4]{Muhammad Faizan Khan}
\ead{20220246@m.scnu.edu.cn}

\author[label5]{Muhammad Muqtasid}
\ead{mmuqtasid875@gmail.com}

\author[label6]{Moamin Abughazala}
\ead{moamin.abughazala1@univaq.it, m.abughazaleh@najah.edu}

\author[label7]{Adnan Akhunzada}
\ead{adnan.akhunzada@udst.edu.qa}


\affiliation[label1]{
    organization={School of Computing, University of Buckingham},
    city={Buckingham},
    postcode={MK181EG},
    country={United Kingdom}
}

    \affiliation[label2]{
    organization={Department of Computer Science Sir Syed CASE Institute of Technology, Islamabad},
    postcode={44000},
    country={Pakistan}
}
    
\affiliation[label3]{
    organization={Department of Data Science and Artificial Intelligence, Faculty of
Information Technology, Al-Ahliyya Amman University, Amman},
    postcode={19328},
    country={Jordan}
}

\affiliation[label4]{
    organization={Aberdeen Institute of Data Science and Artificial Intelligence, South China Normal University, Guangzhou},
    country={China}
}
\affiliation[label5]{
    organization={Department of Computer Science, Faculty of Computing Sciences, Lahore Leads University},
    city={Lahore},
    country={Pakistan}}
\affiliation[label6]{
    organization={Department of Information Engineering, Computer Science and Mathematics, University of L'Aquila},
    city={L'Aquila},
    postcode={67100},
    country={Italy}
}
\affiliation[label7]{
    organization={Department of Data and Cybersecurity, University of Doha for Science
and Technology},
    city={Doha},
    postcode={24449},
    country={Qatar}
}



\begin{abstract}
\noindent
Artificial intelligence (AI) has several transformative effects on healthcare, like to facilitate clinicians in disease detection and to manage patient care
with enhanced precision. Due to the increasing complexity of medical  data and the growing volume of diagnostic images, the substantial computational power needed to process such information creates both opportunities and challenges. To accomplish this objective, numerous methodologies have been proposed to integrate AI into disease detection systems. All these methodologies consider different aspects of clinical diagnosis, considering key parameters such as the type of AI model employed, the nature of input data, and the population demographics being analysed. In this paper, we review different state-of-the-art techniques which have been proposed for various areas of disease detection, specifically focusing on bias identification and mitigation. The proposed solutions are distinguished based on the type of input data they utilize, the architecture of model, and type of bias (algorithmic bias, demographic bias, model bias, and deployment bias). The time frame chosen to filter out relevant studies is between 2015 and 2026. Eventually, we review 23 related articles out of 64,389 analyzed from seven different digital libraries. The obtained results guide us to reach a conclusion with respect to state-of-the-art selected studies.

Furthermore, these results reveal significant research gaps that can assist researchers in addressing AI-related bias in medical diagnosis. By prioritizing explainability, data diversity, and fairness in algorithmic design, as well as equitable access for affected patient populations, they can utilize available healthcare resources more effectively.


\end{abstract}




\end{frontmatter}


\section{Introduction}

Artificial intelligence (AI) is changing healthcare in ways we could hardly anticipated a few years back. Now it is helping doctors to diagnose diseases earlier and more accurately \cite{Smith2020AImedicine} \cite{Lee2021AIHealthcare}. AI systems understand complex and large-scale datasets with help of machine learning (ML) and deep learning (DL) models for identification as well pattern learning. These datasets consist on medical records, imaging scans, genetic information, and even patterns from internet searches
 \cite{brown2019cmaj} \cite{sahu2022precision}. AI models are helpful in making medical decisions, as they provide better diagnostic results compared to humans that can commit a mistake and to reduce chance of errors \cite{rajkomar2019mlmedicine}. Early disease detection is particularly important for serious illnesses such as cancer, Alzheimer’s disease, or infectious diseases. When diagnosed early, treatments work better and patient outcomes improve \cite{Ahmed2020} \cite{Lopez2021}. AI can also extend specialized care to remote or underserved areas through telemedicine and real-time monitoring \cite{Singh2022}. Beyond diagnostics, it can help track disease trends, predict progression, and guide public health decisions. So, AI makes healthcare faster, more precise, and more proactive. \\

AI is powerful, but it is not perfect due to one major problem that is called bias. These systems learn from data, and if the data is incomplete or skewed, the results can be unfair \cite{topol2019}. For example, an algorithm trained mostly on patients from one community may perform well for them but poorly for others. The same issue arises when hospital data mostly comes from wealthy cities. Models built on such data may miss patterns common in rural or low-income areas \cite{obermeyer2019}. These gaps are not abstract they can mean delayed diagnoses, slower treatment, or worse outcomes for those already at risk \cite{mehrabi2021}. Bias can occur at any stage: during data collection, while training the model, or even when the tool is used in practice. That is why bias is not just a technical issue; it is also an ethical one. If patients cannot trust AI to treat them fairly, its potential in healthcare will never be fully realized \cite{char2018}. \\

Real-life examples show how bias affects disease detection. In cancer care, some imaging models detect tumors more accurately on lighter skin than on darker skin, creating unequal results \cite{keesara2020}. During the COVID-19 pandemic, AI tools trained on data from developed countries often failed when applied in low resource regions. These systems simply hadn’t seen enough diverse cases \cite{wynants2020}. Similar challenges appear in Alzheimer’s research, where models built on data from high-income countries may miss lifestyle or cultural differences that affect symptoms elsewhere \cite{sahu2022precision}. Even detection of infectious diesease can be problematic: models may fail in rural or crowded areas where data are scarce, leaving blind spots in outbreak prediction \cite{mehrabi2021}. These examples show that bias is real and tangible. It affects patients’ lives, and those already at the margins of healthcare face the highest risks \cite{Mudassir2021NaturalDisasters}. \\

Fixing bias in AI is not just about better algorithms, in fact it is all about quality of data. Models need information from people of all ages, genders, ethnicities, and regions \cite{mehrabi2021}. Algorithms must be transparent, and they should be tested regularly across diverse populations \cite{wynants2020}. Continuous monitoring is also essential, because healthcare systems and patient needs change over time. Collaboration is equally important where computer scientists, clinicians, and policymakers must work together. Each group contributes knowledge to spot hidden biases and make systems fairer   \cite{jack2018}. There is also an ethical responsibility where AI should reduce inequalities, not worsen them. Clear guidelines and regulations can help ensure that these systems are safe, fair, and trusted \cite{liu2019}.\\

Bias in AI for disease detection is a growing concern. Many studies highlight AI’s benefits, but fewer examine its risks in a structured way \cite{Mehta2019aihealth}. There is a need to map the research that has been done and identify the gaps that remain. A systematic mapping study can provide this overview. It can show which types of bias are most common, which methods are used to detect and reduce them, and which diseases have been studied most \cite{mckinney2020}. The goal of this study is to organize the existing evidence and point toward future directions. Ultimately, the aim is to support AI systems that are not only accurate but also fair and reliable for all patients \cite{obermeyer2019}. \\

This paper presents recent insights on bias in AI for disease detection. It explains both the strengths and the weaknesses of current approaches. The findings of this mapping study will support future research and also encourage new ideas in this important field. The study also shed light on unresolved research gaps, offering opportunities for future work. \\

The major contributions of our work include: 
\begin{itemize}
\item An overview of the state of the art on bias in AI disease detection, based on studies published between 2015 and 2026 in major digital libraries.
\item  A review of research trends in bias detection and mitigation over this period.
\item A summary of the limitations found in primary studies and the open issues that still need attention.
\end{itemize}
The rest of the paper is structured as follows: Section 2 presents the background, Section 3 explains the systematic mapping study process,  Section 4 presents the results of systematic mapping study, Section 5 discusses potential research area, Section 6 presents threats to validity, Section 7 is about discussion and Section 8 concludes the paper.

\section{Background}
Algorithmic bias in healthcare remains a major concern. Race-based corrections for glomerular filtration rate and pulmonary function continue to affect clinical decisions \cite{ImplicitBias2017}. Addressing bias in AI is even harder, as model complexity and varying user literacy can hide discriminatory outcomes. Concealed discrimination is not new in healthcare and often occurs regardless of intent. Implicit bias, for example, delays melanoma diagnoses in dark-skinned patients, resulting in worse outcomes \cite{ImplicitBias2017}. It also contributes to racial disparities in pain medication administration \cite{RacialDisparitiesPain2016}. At the population level, mechanisms such as the “submerged state” reinforce inequities. Here, concealed welfare programs like tax credits perpetuate the racial wealth gap \cite{SubmergedState2014}. During COVID-19, CARES Act funds were allocated based on lost hospital revenue, underfunding hospitals in communities of color \cite{CARESAct2020}. These cases highlight the risk of bias becoming “submerged” within AI systems.

AI adoption in healthcare has grown rapidly. Early studies aimed to improve predictive performance using deep learning and ensemble techniques \cite{Esteva2019} \cite{Mudassir2022REPAIR}. Recent work focuses on fairness, explainability, and trustworthiness \cite{Chouldechova2020}. High accuracy does not guarantee ethical outcomes. Bias in AI often reflects societal inequities in clinical data \cite{Gianfrancesco2018}. \cite{Binns2018} showed that a risk prediction algorithm underestimated the needs of Black patients by using healthcare costs as a proxy \cite{SocioeconomicAI2019}.

Frameworks exist to understand and reduce bias. \cite{Binns2018} categorizes bias as pre-existing, technical, or emergent (Bias Taxonomy). Quantitative fairness measures include statistical parity, equalized odds, and calibration \cite{Hardt2016} \cite{Kleinberg2017}. Healthcare poses unique challenges because decisions can be life-or-death \cite{SocioeconomicAI2019} \cite{Mudassir2025SustainableReconstruction}. Research spans EHRs, imaging, genomics, and clinical triage. \cite{Rajkomar2018} recommended including social determinants of health to correct bias in predictive models \cite{Rajkomar2018}. Despite progress, gaps remain between fairness measures and real-world implementation \cite{Mudassir2025REPAIR}.

Evidence shows disparities across race, gender, socioeconomic status, and geography. Black patients with complex needs are often sicker than White patients at similar risk scores, yet cost-focused algorithms fail to capture this (Risk Prediction Disparities, 2019). Gender bias exists in medical imaging datasets \cite{GenderBias2020}. AI for skin cancer and diabetic retinopathy detection underperforms for underrepresented or lower-income groups \cite{SkinCancerAI2019}\cite{DiabeticRetinopathy2018}. Clinical decision support systems for sepsis in COVID-19 patients also show bias, with poorer patients receiving fewer medications and tests \cite{COVID19Sepsis2020}\cite{Malik2026HyMoBYSwin}\cite{Khan2026LLMNotifications}. These biases mirror those in broader algorithmic systems affecting jobs, housing, and credit \cite{HiringBias2018}\cite{SocioeconomicAI2019}. Overall, AI in healthcare can reinforce inequities unless bias is actively addressed.
\section{Systematic Mapping Study Protocol}
\label{sec:smp}
This section outlines the methodology followed in carrying out the systematic mapping study, based on the guidelines set out by Petersen et al.\cite{petersen2015guidelines}. Figure 2 summarizes the overall workflow. In the figure, rounded rectangles represent processes, while grey rectangles represent the outcomes of each process. The overall process comprises six stages: formulating the research questions, carrying out the search, screening the retrieved papers, extracting keywords and themes, and performing data extraction and mapping. Each step yields a specific outcome, which ultimately forms the basis of our mapping study. 

When conducting our research, it is essential to ensure that the platforms used are authenticated and well-founded. Accordingly, we utilise a range of trusted databases, digital libraries, and platforms, including IEEExplore, Springer, Science Direct, MDPI, Scopus, ACM and Web of Sciences. Figure \ref{fig:prisma} illustrating the breakdown of search query results, followed by the materials retained after screening. In the following sections, we describe each step of the workflow and its corresponding outcome in detail.

\label{sec:mapping_study_protocol}

\subsection{Research objectives and questions}
\subsubsection{Setting the context}

For our research, we employ various platforms to execute subject-specific search queries. Subsequently, filter the relevant materials by applying exclusion and inclusion criteria to ensure that the gathered material is as pertinent as possible. In this way are able to compile a range of different sources including articles, journals, conference papers, and surveys. By applying different criteria, we retain sources most relevant to our research topic.  
\subsubsection{Research questions }
\label{subsec:RQ}
To help narrow down the scope of research, we need to break down the different areas we want to investigate. This allows us to conduct research that is accurate enough for us to reach justified conclusions. \\

\noindent\textbf{RQ1:} What kind of diseases have been addressed by AI? \\

\noindent\textbf{Aim:} This research question aims to analyse the different healthcare and medical domains examined by the various sources. Additionally, this question targets the different models and the respective roles they play in diagnosing various types of diseases and infections. Examples include eye diseases, cardiovascular diseases, lung diseases, skin diseases, and diabetes.\\

\noindent\textbf{RQ2:} What approaches have the publications used to address the issue of bias?\\

\noindent\textbf{Aim:} This question serves as a method for deconstructing the different ways in which the sources have conducted their research. By doing so, we are able to ensure that the information presented in the sources is not only justified with evidence but also accurately supported by other qualified professionals and publications.\\

\noindent\textbf{RQ3:} What parameters have been used in the proposed approaches? \\

\noindent\textbf{Aim:} To maintain relevancy, we need to consider what parameters are used in the filtered material. For example, we identify which parameters have been used and which of them can introduce bias (e.g., feature selection, threshold values, or sampling ratios).\\

\noindent\textbf{RQ4:} What are the reported challenges in disease detections regarding the use of AI?\\

\noindent\textbf{Aim:} By ensuring that the filtered sources include the limitations associated with the use of AI in disease detection, we are able to confirm that the material incorporates the aspect of bias in AI-based disease detection. This further enables us to gather relevant points to support justified claims regarding how bias affects AI in disease detection and diagnosis.\\

\noindent\textbf{RQ5:}
What are top published venues? 
	What are the research trends in last 10 years? 
What kind of expertise required? \\

\noindent\textbf{Aim:}  To identify key publication venues, research trends over the last ten years, and required expertise. These factors help us detect anomalies and observe how bias in AI has fluctuated over time in disease detection.\\

\noindent\textbf{RQ6:} How can the research contributions be classified? \\

\noindent\textbf{Aim:} This research question helps analyse how the filtered material can be distinguished from previous publications and research. Factors such as experiments, testing, and synthetic data are taken into account when differentiating between the different types of sources and the ways in which they can be utilised in our research.\\

\noindent\textbf{RQ7:} What empirical methods have been applied to validate the research contributions? \\

\noindent\textbf{Aim:} Here we focuses on ensuring data diversity and validity in the sources, guaranteeing that the filtered material includes a range of methods for justifying the statements being made. Several sources incorporate inclusion and exclusion criteria, as well as research questions and topic-specific queries. This ensures that the material used to conduct the research in the given sources is appropriately selected and filtered  thereby reducing the risk of misinformation.\\

\subsubsection{Literature Review}
\label{subsubsec:lr}
\begin{figure}[th]
\centering
\includegraphics[width=\textwidth]{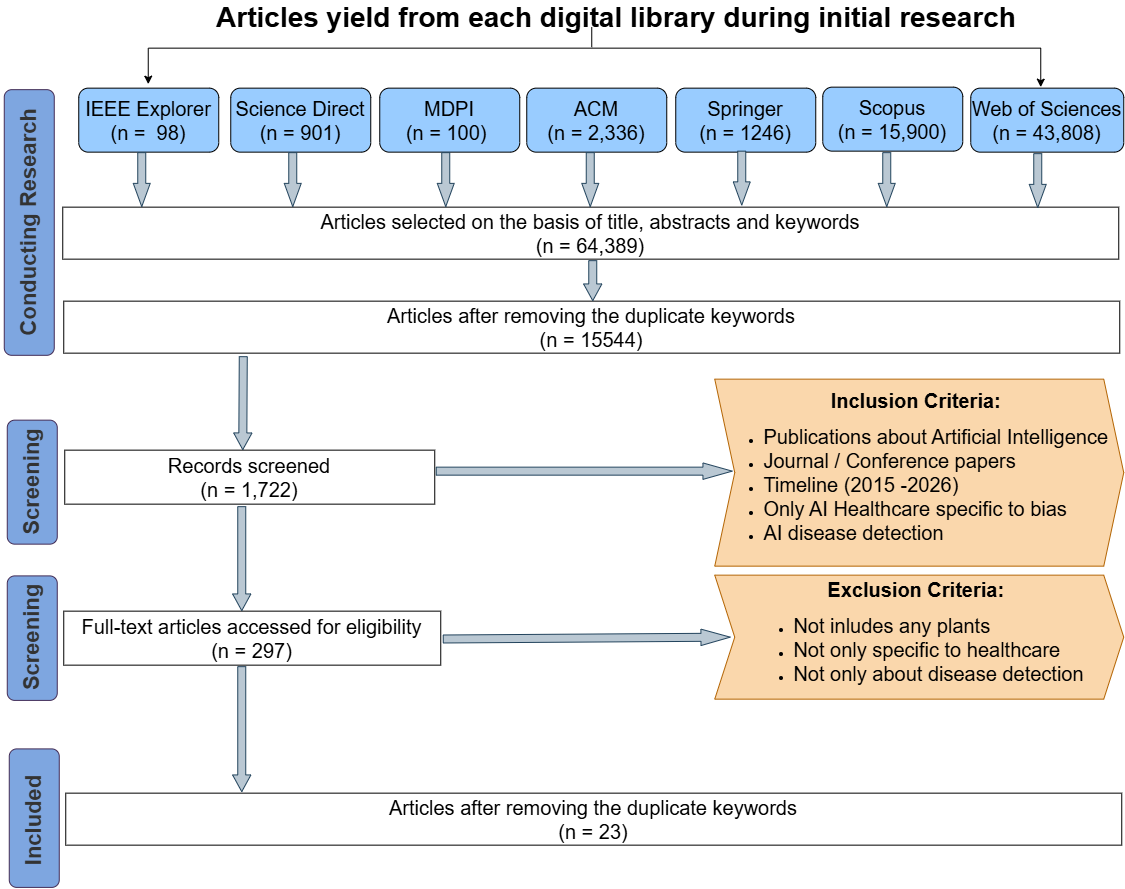}
\caption{PRISMA flow diagram performed in our study}
\label{fig:prisma}
\end{figure}

We adopt the Preferred Reporting Items for Systematic Reviews and Meta-Analyses (PRISMA) framework for literature identification and review. Figure \ref{fig:prisma} illustrates the core PRISMA stages (represented in the left-most blue vertical columns) along with the number of papers involved at each step.

The Conducting Research phase concerns the development of research protocols through the identification of keywords and bibliographic databases used to carry out the search. The Screening phase involves establishing inclusion and exclusion criteria, as well as a defined time frame, for filtering relevant articles. Finally, the Included phase presents the primary articles selected following the application of the screening criteria. Each phase of the PRISMA workflow is described in detail below.
\subsubsection*{Conducting Research} \label{subsubsec:cr}
The PICOC (Population, Intervention, Comparison, Outcome, Context) ~\cite{petticrew2008systematic} criteria are applied to formulate search strings derived from the research questions.

\begin{itemize}
    \item \textbf{Population:}  Refers to all individuals affected by or subject to disease detection using AI systems.
    
    \item \textbf{Intervention:} Denotes the approaches employed to address bias-related issues, such as the algorithms, explainability models, or fairness techniques applied in AI-based disease detection.
    
    \item \textbf{Comparison:} Compares different approaches from the Intervention step, evaluating alternative strategies from a qualitative rather than empirical perspective.
    
    \item \textbf{Outcome:} Focuses on key factors within the examined methodology, including the level of bias, fairness, accuracy across demographic groups, and interpretability of AI models in disease detection.
    
    \item \textbf{Context:} Encompasses works originating from both industry and academia.
\end{itemize}

To identify the most relevant studies published over the last twelve years (2015–2026), we conduct an automated search across seven digital libraries: IEEE Explorer, \footnote{\url{https://ieeexplore.ieee.org/Xplore/home.jsp}} ACM,\footnote{\url{https://dl.acm.org/}} Science Direct,\footnote{\url{https://www.sciencedirect.com/}}
 Springer Link,\footnote{\url{https://link.springer.com/}} Web of Sciences,\footnote{\url{https://apps.webofknowledge.com/}}  Scopus,\footnote{\url{https://www.scopus.com/}} and MDPI.\footnote{\url{https://www.mdpi.com/}}. These libraries provide broad coverage of the academic literature and are selected based on criteria reported in prior work. A set of keywords and associated synonyms is identified, from which search strings are constructed using Boolean operators (AND and OR). The complete query strings are presented in Table \ref{tab:Databases}.

The Quasi-Gold Standard is employed to identify the most relevant primary studies within the specified time frame (2015–2026) across both conferences and journals \cite{zhang2011identifying}. IEEE and ACM Digital Libraries are selected due to their coverage of leading international journals and prominent conferences focused on data science and disease detection. ScienceDirect and SpringerLink are included as comprehensive digital libraries offering extensive scientific collections and indexed international journals. The Web of Science is considered given its standing as one of the oldest citation indexes in the sciences, encompassing scholarly journals, conference proceedings, and data compilations \cite{10.1162/qss_a_00018}.

The Quasi-Gold Standard, is grounded in two primary criteria: the venue and the time period \cite{zhang2011identifying}. Applying these criteria, we identify relevant publications across the seven specified databases through a combination of automated and manual searches. The Quasi-Gold Standard further serves as a useful instrument for evaluating search strategies designed to locate pertinent literature within a defined time frame.

With the aid of the Quasi-Gold Standard (QGS), we compile a collection of studies from related venues, including domain-specific conferences and journals recognised within the time span of 2015 to 2026.

A notable advantage of employing QGS lies in its capacity to efficiently filter relevant studies by focusing exclusively on article titles and abstracts. Conventional manual and automated search methods can prove considerably time-consuming when sifting through large volumes of data and narrowing down publication venues; QGS, however, streamlines this process substantially. Following the identification of relevant studies through QGS, we apply the inclusion and exclusion criteria outlined in the corresponding subsection. As a result, we select 23 primary studies from the available venues.

In addition, we apply the backwards snowballing technique to minimise the risk of overlooking any relevant primary studies published from 2015 onwards \cite{wohlin2014guidelines} \cite{Mudassir2021SocialBasedCity}. This technique involves systematically reviewing the reference lists cited within relevant papers, thereby increasing the likelihood of uncovering additional pertinent studies. Throughout this process, we follow the guidelines established by Wohlin.
\begin{longtable}{p{0.22\linewidth}|p{0.73\linewidth}}
\caption{Search strings in Databases. \label{tab:Databases}} \\
\hline
\textbf{Database} & \textbf{Search String} \\
\hline
\endfirsthead

\multicolumn{2}{c}{\textit{Table \thetable\ (Continued)}} \\
\hline
\textbf{Database} & \textbf{Search String} \\
\hline
\endhead

\hline

\endfoot

\hline
\endlastfoot

\textbf{IEEE Xplore} & 
\parbox[t]{\linewidth}{
(("Artificial Intelligence" OR "Machine Learning" OR "Deep Learning" OR "Neural Network*" OR "AI-based model*") AND 
("Algorithmic Bias" OR "Data Bias" OR "Fairness" OR "Bias Mitigation" OR "Ethical AI" OR "Algorithmic Fairness") AND 
("Disease Detection" OR "Disease Diagnosis" OR "Medical Diagnosis" OR "Clinical Decision Support System*" OR "Medical Imaging")) AND NOT 
("Finance" OR "Stock Market" OR "Agriculture" OR "Recommendation System*" OR "Traffic Detection")
} \\
\hline

\textbf{ACM} & 
\parbox[t]{\linewidth}{
(("Artificial Intelligence" OR "Machine Learning" OR "Deep Learning" OR "Neural Network*" OR "AI-based system*") AND 
("Bias" OR "Algorithmic Bias" OR "Data Bias" OR "Algorithmic Fairness" OR "Discrimination" OR "Ethical AI") AND 
("Disease Detection" OR "Disease Diagnosis" OR "Medical Diagnosis" OR "Clinical Decision Support System*" OR "Medical Imaging" OR "Healthcare AI")) AND NOT 
("Agriculture" OR "Finance" OR "Stock Market" OR "Recommendation System*" OR "Fraud Detection" OR "Traffic Detection" OR "Autonomous Vehicles")
} \\
\hline

\textbf{Science Direct} & 
\parbox[t]{\linewidth}{
(("Artificial Intelligence" OR "Machine Learning" OR "Deep Learning") AND 
("Bias" OR "Algorithmic Bias" OR "Data Bias" OR "Fairness") AND 
("Disease Detection" OR "Disease Diagnosis" OR "Medical Diagnosis" OR "Medical Imaging")) AND NOT 
("Finance" OR "Stock Market" OR "Agriculture")
} \\
\hline

\textbf{Springer Link} & 
\parbox[t]{\linewidth}{
(("Artificial Intelligence" OR "Machine Learning" OR "Deep Learning" OR "Neural Network*" OR "AI-based system*") AND 
("Bias" OR "Algorithmic Bias" OR "Data Bias" OR "Algorithmic Fairness" OR "Ethical AI" OR "Discrimination") AND 
("Disease Detection" OR "Disease Diagnosis" OR "Medical Diagnosis" OR "Clinical Decision Support System*" OR "Medical Imaging" OR "Healthcare AI")) AND NOT 
("Finance" OR "Stock Market" OR "Agriculture" OR "Recommendation System*" OR "Traffic Detection" OR "Autonomous Vehicles")
} \\
\hline

\textbf{Scopus} & 
\parbox[t]{\linewidth}{
(("Artificial Intelligence" OR "Machine Learning" OR "Deep Learning" OR "Neural Network*" OR "AI-based system*") AND 
("Bias" OR "Algorithmic Bias" OR "Data Bias" OR "Algorithmic Fairness" OR "Ethical AI" OR "Discrimination") AND 
("Disease Detection" OR "Disease Diagnosis" OR "Medical Diagnosis" OR "Clinical Decision Support System*" OR "Medical Imaging" OR "Healthcare AI")) AND NOT 
("Finance" OR "Stock Market" OR "Agriculture" OR "Recommendation System*" OR "Traffic Detection" OR "Autonomous Vehicles")
} \\
\hline

\textbf{MDPI} & 
\parbox[t]{\linewidth}{
(("Artificial Intelligence" OR "Machine Learning" OR "Deep Learning" OR "Neural Network*" OR "AI-based system*") AND 
("Bias" OR "Algorithmic Bias" OR "Data Bias" OR "Algorithmic Fairness" OR "Ethical AI" OR "Discrimination") AND 
("Disease Detection" OR "Disease Diagnosis" OR "Medical Diagnosis" OR "Clinical Decision Support System*" OR "Medical Imaging" OR "Healthcare AI")) AND NOT 
("Finance" OR "Stock Market" OR "Agriculture" OR "Recommendation System*" OR "Traffic Detection" OR "Autonomous Vehicles")
} \\

\end{longtable}

\subsubsection{Screening of Papers}

In this section, the inclusion and exclusion criteria are defined for studies retrieved through both automated and manual searches. Two authors independently screened the titles, abstracts, and keywords of the retrieved records against the predefined inclusion and exclusion criteria. Records considered potentially eligible proceeded to full-text assessment. The same two authors independently assessed the full texts. Disagreements at either stage were resolved through discussion and consensus. The following inclusion criteria is applied:

\begin{itemize}
\item  I1: Studies addressing research methods and findings within the considered research domain.
\item  I2: Studies that have undergone a peer-review process and have been published in recognized venues, including journals, conference proceedings, and workshop proceedings.
\item I3: Studies concerned with Artificial Intelligence biasses.
\item I4: Studies published between 2015 and 2026.
\item I5: Studies written in the English language.

\end{itemize}

Studies meeting at least one of the following exclusion criteria are omitted from the review:

\begin{itemize}
\item E1: Studies not focused on AI biasses.
\item E2: Grey literature, including working papers, white papers, short papers, or presentations published in the form of panel discussions.
\item E3: Studies solely proposing guidelines or recommendations to use AI in medical sciences.
\item E4: Studies that have not undergone peer review.
\item E5: Studies not written in English.
\item E6: Duplicate studies published across different venues at various stages of their development.
\end{itemize}

The inclusion and exclusion criteria are applied to all relevant studies identified across the digital libraries to assess their eligibility for the research, encompassing a validated time span of 2015 to 2026. During the full-text review stage, further articles that fail to satisfy the defined criteria are excluded. For the remaining articles, a snowballing technique is subsequently employed, through which three additional relevant studies are identified.

To ensure the reliability of the studies included within the systematic mapping, Fleiss’ Kappa is utilized as a statistical measure of inter-rater agreement. This method evaluates the level of consensus among a fixed number of raters classifying items according to their nature, producing a score ranging from 0, representing poor agreement, to 1, representing full agreement. To assess the reliability of the eligibility criteria, four independent raters evaluated the same randomly selected sample of 10 records, comprising five studies included in the final mapping and five records excluded during screening. The raters independently classified each record as ‘include’ or ‘exclude’ using the predefined eligibility criteria. These raters were not involved in the main screening process; their role was limited to independently validating the consistency of the study-selection decisions. Fleiss’ kappa was used because four raters evaluated the same categorical decisions. The resulting Fleiss’ kappa value of 0.94 indicated almost perfect agreement. The computed Fleiss’ Kappa value of 0.94 demonstrates a high degree of agreement among the raters.

 \begin{figure}[th]
\centering
\includegraphics[width=.7\textwidth]{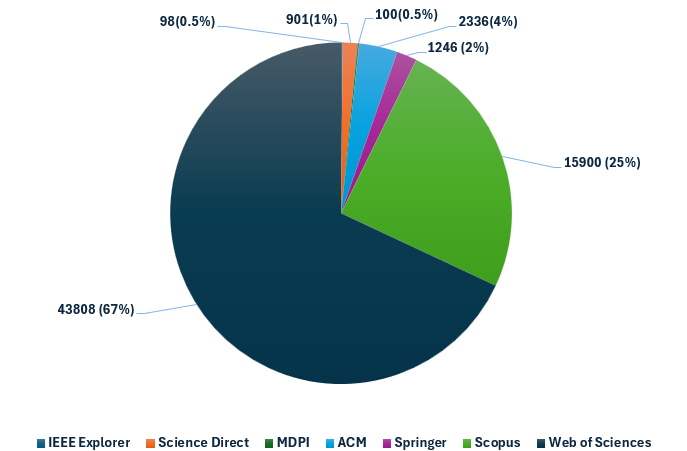}\caption{Primary studies percentage in each database}\label{fig:dbp}
\end{figure} 
Figure \ref{fig:dbp} further illustrates the percentage distribution of primary studies across seven digital libraries. As reflected in the distribution,Web of sciences accounts for the largest proportion at 67\%, followed by Scopus at 25\%, ACM at 4\%, Springer at 2\%, IEEE Xplore and Science Direct at 0.5\% each of the total selected primary studies.
\subsubsection*{Keywords and Themes} \label{subsubsec:kwat}
Following a thorough review process and the systematic application of inclusion and exclusion criteria in accordance with the PRISMA approach, a total of 23 articles are selected for inclusion in this study.

Figure 5 presents the distribution of primary studies across each digital library. Web of Sciences contributes the greatest number of primary studies, accounting for 6 articles. Scopus follows with 5 articles, while ACM includes 4 articles. The Springer digital library comprises 3 articles, and Science Direct consist on 2 articles while MDPI and IEEE Explore  each contain 1 article. It is further noted that one article is indexed across more than one libraries (IEEE Xplore + Scopus). 
\begin{figure}[th]
\centering
\includegraphics[width=.7\textwidth]{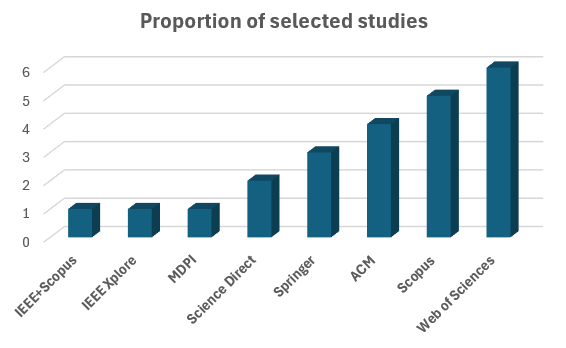}\caption{Selected primary studies}\label{fig:ps}\end{figure}

\paragraph{Quality Assessment:} Following the selection of 23 relevant studies, a quality assessment is conducted drawing upon established quality assessment models from prior literature \cite{adams2017shades},  \cite{yasin2012quality}, \cite{kitchenham2009systematic} and  \cite{tyndall2018aacods}. Based on the recommendations provided within these models, a quality assessment checklist created, as presented in Table \ref{tab:checklist}. This checklist encompasses six distinct criteria and incorporates both subjective and objective questions designed to verify the validity of the sources, evaluate their suitability, and ensure they are free from bias.

Then checklist subsequently applied to assess the quality of the selected literature. Through the implementation of specific selection criteria, all irrelevant studies are effectively excluded, thereby minimising the effort required for the quality assessment process \cite{garousi2019guidelines}. These criteria are grounded in factors such as the authority of the producer, methodology, scope, objectivity, novelty, and impact \cite{Mudassir2020PhysicalReconstruction}.

The same selection criteria may further be applied to other research methods, including surveys, case studies, and experiments, to support broader quality evaluation. A quality assessment checklist specifically developed for case studies has been proposed in prior work and may equally be applied to case studies within formal literature.

\subsubsection*{Data Extraction \& Mapping Process}
To extract data from the selected primary studies and address the research questions defined in the corresponding subsection \ref{subsec:RQ}, Table \ref{tab:data-extraction} is constructed to incorporate the key data items related to the research questions. In line with established practice, one researcher conducted the data extraction while another reviewed and verified the process. This two-author approach to verification is a widely adopted practice in systematic mapping studies \cite{petticrew2008systematic}. The defined data extraction strategy ensures that consistent criteria are applied uniformly across all selected papers, including their respective classifications.
\begin{table}[H]
\caption{Quality assessment checklist of relevant literature for mapping study on AI bias in disease detection.
\label{tab:checklist}}
\begin{tabular}{p{0.29\linewidth}p{0.65\linewidth}}
\hline
\hline
\textbf{Criteria} & \textbf{Questions} \\
\hline
\hline
Producer Authority 
& • Is the publishing institution or platform reputable? E.g., \textit{Journal of Biomedical Informatics}, \textit{Nature Medicine} \newline
- Is the author affiliated with a recognised research or medical institution? \newline
- Has the author published prior work in AI, healthcare, or disease detection? \newline
- Does the author possess relevant expertise in this area? (e.g., AI researcher, clinical data scientist, or medical professional) \\
\hline
Methodology 
& • Does the proposed methodology have a clearly stated aim related to AI-based disease detection? \newline
- Is the proposed approach supported by authentic and peer-reviewed references? \newline
- Are the limitations of the proposed methodology clearly acknowledged? \newline
- Is the work grounded in specific research questions, with all questions addressed effectively? \newline
- Is the proposed methodology validated through a real-world clinical case study or dataset? \\
\hline
Scope 
& • Does this work relate to the domain of AI bias in healthcare or disease detection? \\
\hline
Objectivity 
& • Is the work balanced in its presentation and clearly articulated? \newline
- Does the problem statement clearly define the research objective? \newline
- Is this work associated with or influenced by a particular vendor or commercial platform? \newline
- Are the conclusions of this work supported by empirical data or experimental evidence? \\
\hline
Novelty 
& • Does this work introduce a new idea, approach, or insight regarding bias in AI-driven disease detection? \\
\hline
Impact 
& • Assess the impact of the considered study with respect to the following criteria: \newline
- Number of paper citations \newline
- Number of paper backlinks \newline
- Number of paper views or reads \\
\hline
\end{tabular}
\end{table}
\begin{table}[H]
\centering
\caption{Data Extraction}
\label{tab:data-extraction}
\begin{tabular}{p{3.9cm} p{9cm} c}
\hline

\textbf{Data Item} & \textbf{Value} & \textbf{RQ} \\
\hline
Article ID & Integer & \\
Author Name & Author's name list & \\
Title of Articles & Name of the article & RQ1 \\
Keyword & Keyword study indexing & RQ1 \\
Publication Year & Calendar year & RQ5 \\
Venue & Publication venue name & RQ5 \\
Disease Type & Eye diseases, cardiovascular diseases, lung diseases, skin diseases, diabetes, cancer, etc. & RQ1 \\
AI Models Used & Machine Learning, Deep Learning, CNN, RNN, Decision Trees, Random Forests, SVM, LLMs, NLP & RQ1 \\
Research Approach & Tool-based, solution methodology, case study, systematic review, empirical study & RQ2 \\
Bias Mitigation Strategy & Algorithmic debiasing, adversarial learning, fairness constraints, reweighting, data augmentation & RQ2 \\
Parameter Type & Feature selection, threshold values, sampling ratios, hyper parameters, demographic variables & RQ3 \\
Bias-Inducing Parameters & Imbalanced datasets, missing demographic data, skewed sampling ratios, improper threshold selection & RQ3 \\
Reported Challenge & Data quality issues, algorithmic bias, labeling bias, deployment bias, lack of generalizability & RQ4 \\
Limitation of Study & Small sample size, lack of external validation, dataset homogeneity, computational constraints & RQ4 \\
Contribution Type & Novel algorithm, bias detection framework, fairness evaluation metric, comparative analysis & RQ6 \\
Validation Method & Cross-validation, external validation, ablation studies, fairness audits, statistical testing & RQ7 \\
Data Diversity & Multiple datasets, diverse demographic representation, synthetic data generation & RQ7 \\
Search Strategy & Guidelines followed to select the studies (automated/manual, inclusion/exclusion criteria) & RQ4 \\
Search Type & Manual or automated & RQ4 \\
Research Trend & Year-wise publication count, emerging bias topics, model evolution & RQ5 \\
Domain Expertise & Medical imaging, clinical informatics, computer vision, epidemiology, ethics & RQ5 \\
Classification Scheme & How articles were categorized (by disease, model type, bias type, or methodology) & RQ6 \\
Open Issues & Unresolved challenges in bias mitigation, generalization gaps, lack of standardized fairness metrics & RQ4 \\
\hline
\end{tabular}
\end{table}

\textbf{\textit{Analysis and classification:}} \label{sec:analysis_classification}
The extracted information is elaborated upon and visually presented in the corresponding subsection. During the analysis of the study identification phase, the extracted strategies are classified according to different search approaches, which included the development of the search process, the evaluation procedures, and the inclusion and exclusion criteria. Each category is subsequently assigned relevant themes and sub-themes. The sub-themes are used within the inclusion and exclusion process, also referred to as a priori classification, drawing upon strategies such as resolving disagreements between researchers in order to minimize bias \cite{petersen2011identifying}. Finally, the selected papers tallied according to their respective themes and sub-themes.

A descriptive technique  additionally applied, as illustrated in Figure \ref{fig:fr}, to categories the publication types of the selected studies including book chapters, conference papers, journal articles, and symposium contributions according to their respective venues. Furthermore, a classification scheme grounded in topic-related keywords has established to effectively address specific research questions. These classifications prove particularly valuable in supporting the mapping process.

\section{Systematic Mapping Study Results} \label{sec:msr}

We conduct this mapping study following the procedures outlined in Section \ref{sec:smp} and in alignment with our proposed research questions. The initial search phase retrieves 64,389 records based on filtering criteria such as publication year, author affiliations, source venue, and subject relevance, while applying our predefined inclusion standards. After refining the search by eliminating duplicate keywords and adhering to systematic review protocols \cite{petersen2015guidelines}, we identify an additional 292 records that satisfy the inclusion requirements. Following abstract screening and the application of our selection criteria, a total of 23 primary studies remain for final analysis. Appendix A presents a concise summary of all included studies.

\begin{figure*}[th]
\centering
\includegraphics[width=.8\textwidth]{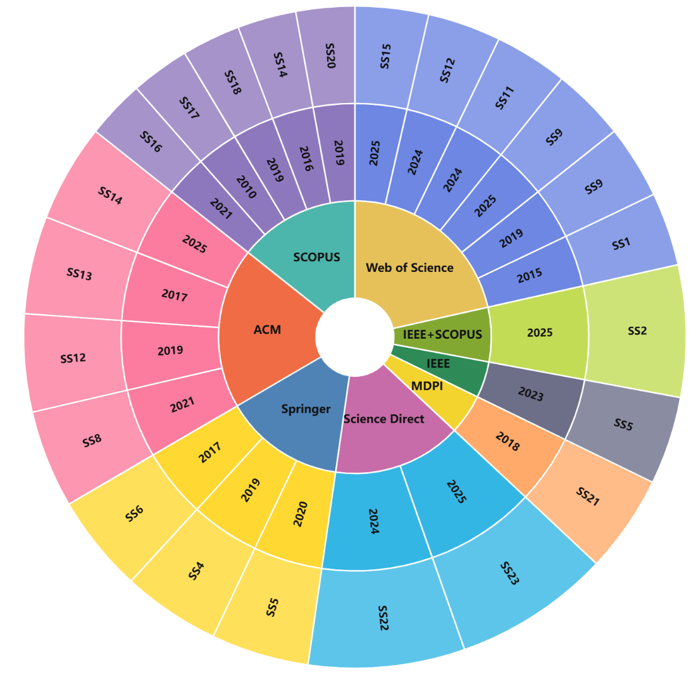}\caption{Distribution of selected studies (SS) used for analysis across publications years }\label{fig:dpt}\end{figure*}

\begin{figure}[th] 
\centering
\includegraphics[width=.7\textwidth]{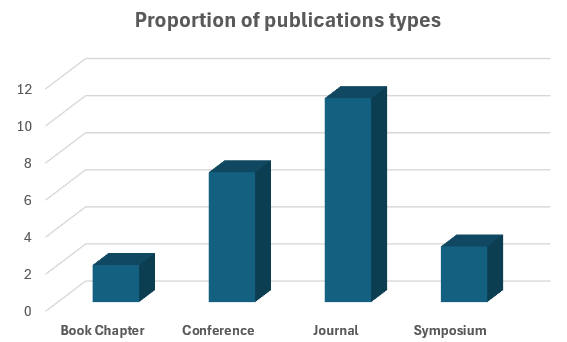}\caption{Distribution of selected studies with respect to publication type}\label{fig:fr}\end{figure}

Figure \ref{fig:dpt} offers a visual summary of the selected primary studies distributed across seven digital repositories. The figure captures the source libraries from which the studies originate including publication year and the outer ring of the pie chart illustrates the proportional breakdown of results derived from the selected studies.


\subsection{Categorization of Primary Studies by Publication Format}

We examine the selected primary studies based on their publication format, which includes journal articles (11 studies), conference proceedings (7 studies), book chapters (2 studies) and symposium (3 studies). Figure \ref{fig:fr} visualizes the distribution across these formats. Journals represent the most frequently used publication channel, comprising roughly 47.8\% of all selected works concerning AI bias in disease detection. In contrast, conference papers account for 30.43\%, symposium proportion is 13\% while book chapters make up the smallest portion at approximately 8.69\%.



\subsection{Classification of Primary Studies by Research Type}

To classify primary studies according to their research nature, we adopt the taxonomy proposed by \cite{wieringa2006requirements} and \cite{durelli2019machine}. These categories remain straightforward and are explained below in Table \ref{tab:dataclassification}.
\begin{table}[ht]
\caption{Studies Classification According To Research Type}
\label{tab:dataclassification}
\begin{tabular}{p{0.55\linewidth}p{0.32\linewidth}}
\hline
\textbf{Research Type} & \textbf{Number of Studies} \\
\hline
Solution Proposal & 3 \\
\hline
Solution Proposal with Evaluation & 3 \\
\hline
Solution Proposal with Validation & 2 \\
\hline
Philosophical papers & 4 \\
\hline
Opinion / Personal Experience & 11 \\
\end{tabular}
\end{table} \\
\textit{Solution Proposal:} Such papers introduce a novel approach to addressing a problem but lack comprehensive validation. Instead, they rely on proof-of-concept demonstrations through illustrative examples, logical arguments, or similar means. Three of our selected studies fall under this category.

\noindent\textit{Solution Proposal with Evaluation:} This category encompasses empirical investigations that assess novel solutions using formal methodologies including hypothesis testing and real-world case study experiments. Three of our primary studies are classified as solution proposals accompanied by evaluation research.

\noindent\textit{Solution Proposal with Validation:} Studies in this group offer preliminary empirical support for proposed solutions through methodologically rigorous procedures such as quasi-experiments, prototyping, mathematical modeling, and case studies. We identify two primary studies in the validation research category.

\noindent\textit{Philosophical Papers:} According to \cite{wieringa2006requirements}, these works introduce new conceptual frameworks or alternative perspectives on existing research. We have found four primary study fits this classification.

\noindent\textit{Opinion Papers:} These articles reflect the authors' personal viewpoints on a research area for instance, identifying deficiencies or advocating alternative techniques. Approximately 49\% of the selected studies lack rigorous empirical validation and instead rely on proof-of-concept illustrations. These eleven works are designated as pure solution proposals. This finding suggests that the existing literature on AI bias in disease detection exhibits relatively limited scientific robustness.

The second category  solution proposals accompanied by evaluation research  comprises three studies. These works contribute high-quality evidence through experiments grounded in real-world case studies, thereby demonstrating the practical applicability of their proposed innovations.

Validation research studies assess the potential contributions of selected works using experimental setups and mathematical analyses. These studies go beyond basic solution proposals by offering deeper methodological scrutiny. The presence of two such studies indicates emerging but still limited scholarly engagement with this domain in recent years \cite{wieringa2006requirements}.

Four philosophical papers appear, suggesting that a modest number of researchers propose entirely new conceptual frameworks in this field. Similarly, eleven opinion or personal experience papers are included, highlighting a strong reliance on subjective insights rather than empirical validation in the existing literature

\subsection{Mapping of Primary Studies by AI Bias Solution Category}

Following data extraction, we classify the selected studies based on the type of AI bias addressed. This classification involves two independent reviewers working in two stages. The first stage consists of systematic mapping, while the second stage establishes four bias domain categories: algorithmic bias, demographic bias, model bias, and deployment bias. Table \ref{tab:ai-bias-classification} summarizes these categories.
\begin{table}[ht]
\centering
\caption{Classification of Primary Studies by AI Bias Domain}
\label{tab:ai-bias-classification}
\begin{tabular}{p{3cm} p{4cm} p{3.3cm} c}
\hline
\textbf{Bias Category} & \textbf{Focus Areas} & \textbf{Associated Studies} & \textbf{Total Count} \\
\hline
Algorithmic Bias & Computational errors, design flaws, pattern recognition issues & SS1, SS2, SS5, SS12, SS15, SS16, SS17 & 7 \\
\hline
Demographic Bias & Gender, racial, age, and socioeconomic disparities & SS3, SS6, SS7, SS8, SS22, SS23 & 6 \\
\hline
Model Bias & Training-related bias, design limitations, black-box opacity & SS4, SS9, SS10, SS13, SS18, SS20, SS21 & 7 \\
\hline
Deployment Bias & Use outside intended context, generalisation failures & SS11, SS14, SS19 & 3 \\
\bottomrule
\end{tabular}
\end{table}

\textit{Algorithmic Bias:} These works examine how computational processes and algorithm design choices can introduce systematic errors into disease detection. We identify 7 such articles, representing 30.4\% of all selected studies.

\textit{Demographic Bias:} This category includes multidisciplinary research on how gender, race, age, and socioeconomic status can contribute to biased outcomes, often mirroring broader societal inequalities. Six studies fall into this group.

\textit{Model Bias:} Focusing on biases stemming from model design and training procedures, this category accounts for seven studies, or also 30.4\% of the total.

\textit{Deployment Bias:} These four studies (13.0\%) investigate bias that arises when AI systems operate outside their originally intended contexts or purposes.

\subsection{Distribution of Studies by Medical Domain}

Through our analysis, we extract information across various study types including solution proposals, empirical investigations, evaluation research, philosophical essays, surveys, and opinion pieces. During data extraction, we record several key attributes, including disease type, AI model employed, bias category, and validation approach. Some studies present information at an abstract level using graphical representations.

\noindent Table \ref{tab:disease-domain} presents the breakdown of selected studies according to the medical domain addressed.

\begin{table}[ht]
\centering
\caption{Distribution of Studies by Disease Domain}
\label{tab:disease-domain}
\begin{tabular}{l c c c}
\hline
\textbf{Medical Domain} & \textbf{Primary Studies} & \textbf{Count} & \textbf{Percentage} \\
\hline
Cardiovascular Diseases & SS2, SS4, SS5, SS14, SS16 & 5 & 21.7\% \\
\hline
Ophthalmology & SS3, SS17, SS18, SS23 & 4 & 17.39\% \\
\hline
Lung Diseases & SS7, SS8, SS19, SS20 & 4 & 17.39\% \\
\hline
Skin Diseases & SS1, SS6, SS12& 3 & 13.0\% \\
\hline
Diabetes &  SS13, SS21 & 2 & 8.69\% \\
\hline
Multiple / General &  SS9, SS10, SS11, SS15, SS22 & 5 & 21.7\% \\
\bottomrule
\end{tabular}
\end{table}

\subsection{Leading Contributors in the Field}

Based on our examination of the selected primary studies, we analyse publication patterns and author productivity to assess research impact within this domain. Several active researchers contribute multiple papers over the past decade, like \cite{petersen2015guidelines}. While publication volumes show notable increases in 2020, 2021, 2024, and again in 2026, our findings indicate that no research group yet comprehensively tackles the problem of bias in black-box AI models using state-of-the-art explainability techniques.

\section{Potential Research Areas}

Based on the findings of this mapping study, we observe that limited research effort has been devoted to this domain. Although some scholars have attempted to address AI bias in disease detection, the overall volume of methodologically rigorous studies remains insufficient. Most available works rely on conceptual frameworks or descriptive analyses rather than empirical investigations. Consequently, several important issues remain underexplored. We identify multiple promising research directions, particularly regarding the application of explainable AI (XAI) techniques to detect and mitigate bias in medical diagnostic systems.

The use of cutting-edge explainability methods in this critical domain offers considerable potential. According to our mapping results, only a small fraction of selected studies employ advanced explainability frameworks to uncover hidden biases in AI models. Many current AI systems operate as black boxes, making it exceedingly difficult for healthcare professionals and decision-makers to trace how diagnostic conclusions are reached. Without transparent and interpretable outputs, biases may remain undetected, potentially leading to misdiagnosis or unequal treatment across patient populations.

Furthermore, public health authorities and clinical decision-makers rarely account for the societal impact and relative benefits that patients experience from AI-assisted diagnosis. Indeed, these societal consequences vary significantly across different demographic groups and clinical settings. Yet such considerations represent essential factors that ought to be integrated into all stages of AI deployment in healthcare from model design and training to validation and post-deployment monitoring.

Given these complexities, explainable AI frameworks offer a viable pathway forward. These approaches can provide clinicians, administrators, and patients with mechanisms to better understand, evaluate, and refine diagnostic algorithms. By making model decisions interpretable, XAI helps identify whether bias arises from training data, algorithmic design, or deployment contexts.

Thus, more empirical and detailed investigations are necessary to examine how explainable AI techniques can effectively detect, measure, and mitigate bias in disease detection systems. Particular attention should be directed toward black-box models operating across diverse medical domains, including cardiology, ophthalmology, pulmonology, dermatology, and endocrinology.

\section{Threat to Validity} \label{sec:validity}

Several factors require careful consideration when evaluating the findings of any systematic mapping study, as these factors may potentially limit the validity of the conclusions drawn. In this section, we discuss potential validity threats to our systematic mapping study with respect to study design, execution, analysis, and interpretation. Specifically, we consider threats such as selection bias in digital library choice, inaccuracies during data extraction, temporal bias in study inclusion, search string construction limitations, and publication bias.

\subsection{Selection Bias in Digital Libraries}

Selection bias refers to potential distortions in statistical analysis arising from the choice of digital libraries specifically, whether selected repositories contain too many or too few relevant studies. We address this threat (at least to some extent) by defining comprehensive inclusion and exclusion criteria tailored to AI bias research in healthcare. Additionally, we select seven widely recognized digital libraries: IEEE Xplore, ACM Digital Library, Springer Link, Science Direct, Scopus, MDPI, and Web of Sciences. This broad coverage minimizes the risk of omitting pertinent studies.

\subsection{Inaccuracy in Data Extraction and Misclassifications}

During information extraction, inaccuracies and misclassifications may occur when reviewers interpret study content differently. To mitigate such discrepancies (at least to some extent), we resolve all disagreements through consensus discussion among all reviewers. Furthermore, we design a standardized data extraction form (Table \ref{tab:data-extraction}) that captures key attributes consistently across all selected studies.

\subsection{Temporal Bias in Primary Study Selection}

Temporal bias arises when the chosen time frame excludes relevant recent publications or overrepresents older works. To address this threat, we restrict our search to articles published between 2015 and 2026. This ten-year window captures the most active period of AI research in healthcare while minimising the chance of missing recent, high-impact contributions.

\subsection{Bias in Search String Definition}

Search string construction introduces potential bias if keywords fail to capture the full scope of relevant literature. To tackle this issue (at least to some extent), we employ multiple search techniques, including Boolean operators (AND, OR), quotation marks for exact phrases, hyphens for compound terms, exclusion operators (NOT), and wildcard asterisks. We also refine our search strings iteratively, removing duplicate keywords and excluding irrelevant domains such as plant disease detection.

\subsection{Publication Bias}

Publication bias refers to the tendency of journals and conferences to publish positive results more frequently than negative or null findings. Additionally, studies reporting negative outcomes often take longer to appear in print and receive fewer citations \cite{Hopewell2007}. To address this threat (at least to some extent), we scan all relevant journals and conference proceedings while deliberately excluding grey literature such as technical reports, unpublished theses, and non-peer-reviewed white papers as these may compromise the validity of our findings.

\subsection{Descriptive Validity}

Descriptive validity concerns the accuracy and objectivity of the information collected. Threats to descriptive validity typically pose greater challenges in qualitative research than in quantitative studies. To address this, we design a standardized data collection form (Table \ref{tab:data-extraction}) to systematically record information related to AI bias in disease detection. We also review our extraction process multiple times to avoid missing any relevant articles. Moreover, this form remains accessible for re-examination whenever needed.

\subsection{Theoretical Validity}

Theoretical validity involves determining whether the collected data appropriately fits the intended analytical framework without selection bias. There always exists a risk of omitting relevant studies for example, \cite{kitchenham2015evidence} conduct two mapping studies on identical topics and arrive at different sets of articles. To handle this threat, we follow PRISMA guidelines \cite{song2021assessment} and employ backward snowball sampling techniques \cite{jalali2012systematic} as described in subsection \ref{subsubsec:lr}.

\subsection{Internal Validity}

Internal validity faces two primary threats: missing relevant studies and researcher bias during paper selection. Mapping studies typically aim to cover a broad range of related articles from multiple databases \cite{kitchenham2015evidence}. In our mapping study, we therefore search seven of the most widely used digital libraries to minimise the chance of overlooking relevant research on AI bias in disease detection. We believe that most pertinent primary studies are included, although we cannot entirely rule out the possibility that some relevant works may have been missed during automated searches across digital repositories. Additionally, we employ a quasi-gold standard approach (subsection \ref{subsubsec:lr}) to validate the completeness of our automated search strings, which are derived from research question keywords.

Researcher bias during paper selection may also lead to inaccuracies in data extraction. To address this, we define clear data inclusion criteria (Table \ref{tab:data-extraction}) for extraction, with all reviewers reaching agreement. Any disagreements arising during extraction are settled through mutual consensus.

\subsection{External Validity}

According to \cite{petersen2013worldviews}, external validity in systematic mapping studies concerns the generalisability and usefulness of findings for other populations or contexts. The main threat, therefore, is whether the selected primary studies adequately represent the subject area. To address this, we follow a standard research process and include all relevant studies regardless of outcome direction, excluding only those not written in English.

Additionally, some primary studies do not provide complete information for the extraction form. In such cases, we infer certain information based on the inclusion criteria during data synthesis. Similarly, some approaches do not explicitly describe their limitations or drawbacks.

\subsection{Conclusion Validity}

Conclusion validity refers to the extent to which we reach reasonable conclusions and establish meaningful relationships between selected studies and current research trends in the field. To handle the first threat, we systematically answer all defined research questions based on the selected primary studies. Regarding the second threat concerning data extraction, classification, and synthesis we perform all activities as a collaborative team. Nevertheless, a residual risk remains that some relevant study may have been missed, as qualitative mapping studies inherently involve interpretation and cannot guarantee absolute completeness.
\section{Discussion}
This study systematically mapped the current state of research on AI bias in disease detection between 2015 and 2026, identifying the major research themes, bias categories, disease domains, research methodologies, and validation approaches. The analysis of 23 primary studies collected from seven digital libraries provides a comprehensive overview of how fairness and bias have been investigated in healthcare AI over the past decade. The findings indicate that although considerable attention has been directed toward improving fairness in AI-assisted diagnosis, substantial challenges remain in developing robust, transparent, and clinically validated bias mitigation frameworks that can be deployed in real-world healthcare settings.

The mapping results demonstrate that recent studies increasingly focus on algorithmic fairness, demographic bias, explainability, and fairness-aware machine learning, reflecting a gradual transition from performance-oriented AI models toward trustworthy and responsible AI systems. Nevertheless, the adoption of advanced bias mitigation techniques remains limited. Only a small number of primary studies employed explainable AI methods or integrated fairness constraints directly into model development, while most research concentrated on identifying bias after model training. This observation suggests that the field is still in the early stages of incorporating fairness-by-design principles into AI-based disease detection and highlights an important opportunity for future research on interpretable and bias-aware learning frameworks.

Furthermore, the distribution of research types indicates limited methodological maturity within the existing literature. A considerable proportion of the selected studies were classified as opinion or philosophical papers, whereas comparatively few studies presented empirically validated solutions using real clinical datasets. Although several solution proposals introduced promising approaches for bias detection and mitigation, the lack of large-scale experimental validation limits their generalizability across different healthcare environments. Moreover, the absence of standardized fairness benchmarks and evaluation protocols makes it difficult to compare the effectiveness of competing approaches. This emphasizes the need for future studies to prioritize reproducible experimentation, external validation, and benchmark datasets representing diverse patient populations.

Another important observation concerns the factors contributing to bias in AI-based disease detection. The identified parameters, including demographic characteristics, data imbalance, feature selection, model architecture, and deployment environment, illustrate that bias is influenced by technical, clinical, and societal factors simultaneously. However, most primary studies examined these factors independently, with limited consideration of their interactions throughout the AI development lifecycle. Such fragmentation restricts the applicability of current solutions in complex clinical environments where multiple sources of bias coexist. A comprehensive framework integrating data diversity, explainability, fairness evaluation, and continuous post-deployment monitoring could provide a more effective strategy for reducing bias while maintaining diagnostic performance.

Finally, the distribution of research across medical domains reveals that cardiovascular diseases, ophthalmology, and respiratory diseases have received the greatest research attention, whereas several clinical specialties remain comparatively underexplored. Similarly, studies involving geographically diverse populations and healthcare systems from low- and middle-income countries remain limited despite the increasing global adoption of AI-assisted diagnosis. This imbalance suggests that future research should place greater emphasis on developing equitable datasets, evaluating models across diverse demographic groups, and strengthening collaboration between computer scientists, clinicians, healthcare professionals, and policy experts. Such interdisciplinary efforts will be essential for ensuring that future AI systems are not only accurate but also fair, transparent, and reliable for clinical decision-making.
\section{Conclusion}
This mapping study identifies and evaluates the existing literature with respect to research topics, publication frequency up to 2026, publication venues, types of AI bias, and disease domains. We believe our mapping study provides a comprehensive and state-of-the-art overview of AI bias in disease detection, which proves highly useful for emerging researchers and healthcare practitioners seeking to understand this challenging field and contribute through bias detection and mitigation strategies.

We answer all defined research questions (RQs) through analysis and results derived from our systematic mapping study. \\

\textbf{RQ1: What kind of diseases have been addressed by AI?}\\

According to our primary studies, we focus exclusively on human disease detection research published over the last ten years (from 2015 to 2026). After thorough examination, we identify five major disease categories addressed in the literature. Table \ref{tab:disease-categories} summarizes these categories along with their corresponding primary studies.

\begin{table}[ht]
\centering
\caption{Disease Categories Addressed in Primary Studies}
\label{tab:disease-categories}
\begin{tabular}{p{4cm} p{7cm}}
\hline
\textbf{Disease Category} & \textbf{Primary Studies} \\
\hline
Cardiovascular Diseases & SS2, SS4, SS5, SS14, SS16, SS19 \\
\hline
Ophthalmology & SS3, SS9, SS10, SS11, SS17, SS18 \\
\hline
Lung Diseases & SS7, SS8, SS20, SS23 \\
\hline
Skin Diseases & SS1, SS6, SS12, SS15\\
\hline
Diabetes & SS13, SS21, SS22 \\
\bottomrule
\end{tabular}
\end{table}

\textbf{RQ2: What approaches have the publications used to address the issue of bias?} \\

The approaches identified in primary studies to address AI bias in disease detection fall into four categories: algorithmic bias detection (examining computational process errors and design flaws), demographic bias analysis (investigating gender, racial, age, and socioeconomic disparities), model bias assessment (evaluating training-related and design limitations), and deployment bias investigation (studying generalization failures and out-of-context usage). Table \ref{tab:bias-approach} groups the primary studies according to these four bias categories.

\begin{table}[ht]
\centering
\caption{Papers Grouped Based on Bias Approach}
\label{tab:bias-approach}
\begin{tabular}{p{3.5cm} p{9.5cm}}
\hline
\textbf{Bias Approach} & \textbf{Primary Studies} \\
\hline
Algorithmic Bias & SS1, SS2,  SS12, SS16, SS17, SS18, SS19, SS20, SS21, SS22 \\
\hline
Demographic Bias & SS5, SS7, SS8, SS10, SS11, SS13, SS14, SS15 \\
\hline
Model Bias & SS3, SS9\\
\hline
Deployment Bias & SS4, SS6, SS23 \\
\bottomrule
\end{tabular}
\end{table}

\textbf{RQ3: What parameters have been used in the proposed approaches?} \\

In selected studies, we analyze all attributes and input parameters employed in the proposed methodologies. We categorize these parameters across four bias domains: algorithmic bias, demographic bias, model bias, and deployment bias. Table \ref{tab:bias-parameters} details the key parameters identified within each bias category. \\

\begin{table}[ht]
\centering
\caption{Key Parameters by Bias Category}
\label{tab:bias-parameters}
\begin{tabular}{l p{8cm}}
\hline
\textbf{Bias Category} & \textbf{Key Parameters} \\
\hline
Algorithmic Bias & Computational processes, algorithm design, pattern recognition, decision-making transparency, explainability scores \\
\hline
Demographic Bias & Gender distribution, racial composition, age range, socioeconomic status, geographical representation \\
\hline
Model Bias & Training data diversity, model architecture, black-box opacity, validation methods, transfer learning applicability \\
\hline
Deployment Bias & Intended use context, real-world generalisation, environmental fit, operational boundaries \\
\bottomrule
\end{tabular}
\end{table}

\textbf{RQ4: What are the reported challenges in disease detection regarding the use of AI?}
\\

Our analysis of primary studies reveals three major challenges in AI-based disease detection, as shown in Table \ref{tab:challenges}. These challenges include data diversity and representativeness, lack of explainability in black-box models, and algorithmic bias perpetuating existing healthcare disparities. \\

\begin{table}[ht]
\centering
\caption{Reported Challenges in AI-Based Disease Detection}
\label{tab:challenges}
\begin{tabular}{|p{4.5cm}|p{8.5cm}|}
\hline
\textbf{Challenge Category} & \textbf{Primary Studies} \\
\hline
Data Diversity and Representativeness & SS1, SS5, SS7, SS10, SS11, SS14, SS15, SS16, SS19, SS21 \\
\hline
Lack of Explainability / Black-Box Opacity &  SS2, SS3, SS9, SS12, SS18, SS22 \\
\hline
Algorithmic Bias and Healthcare Disparities & SS4, SS6, SS8, SS13, SS17, SS20, SS23 \\
\hline
\end{tabular}
\end{table}

\textbf{RQ5: What research factors need to be taken into consideration?} \\

During our search for primary studies, we identify several key factors that require attention when conducting research on AI bias in disease detection. We address these factors through the following sub-questions.

\textbf{\textit{RQ5.1: What are the top published venues?}}

Among the 22 distinct venues identified from selected primary studies, seven conferences and journals emerge as the most popular, containing the highest concentration of relevant publications. These prominent venues include IEEE Xplore, ACM Digital Library, Science Direct, Springer Link, Scopus, MDPI, and Web of Sciences.

\textbf{\textit{RQ5.2: What are the research trends in the last ten years?}}

As it is mentioned research activities fluctuates between 2015 and 2026. Limited scholarly attention appears between 2015 and 2017. A significant increase occurs in 2020, followed by peak activity in 2021. After a modest decline in 2022, publication volumes rebound in 2023 and 2024. The highest research output appears in 2025, reflecting growing concern about AI bias in medical diagnostics.

\textbf{\textit{RQ5.3: What kind of expertise is required?}}

From the primary studies, we understand that AI bias in disease detection intersects multiple disciplines. Effective research in this domain therefore requires expertise in computer science (particularly machine learning, deep learning, and explainable AI), clinical knowledge specific to each disease domain, data science skills for handling diverse medical datasets, and social science perspectives for understanding demographic and ethical implications of algorithmic bias.
\\

\textbf{RQ6: How can the research contributions be classified?}
\\

Table \ref{tab:research-contributions} presents the classification of selected primary studies according to research contribution type, as proposed by [101] and [23]. We distinguish contributions based on methodological approach, validation strategy, and empirical grounding.
\begin{table}[h]
\centering
\caption{Classification of Research Contributions}
\label{tab:research-contributions}
\begin{tabular}{|p{3.5cm}|p{3.5cm}|p{8cm}|}
\hline
\textbf{Contribution Type} & \textbf{Primary Studies} & \textbf{Description} \\
\hline
Solution Proposal & SS6, SS9, SS13, SS15, SS17, SS22 & Introduces novel bias detection or mitigation approach with proof-of-concept validation \\
\hline
Evaluation Research & SS2, SS4, SS5, SS8, SS11, SS14, SS18, SS20, SS21 & Empirically evaluates existing solutions using real-world case studies or formal experiments \\
\hline
Philosophical / Conceptual & SS1, SS3, SS7, SS10, SS16, SS19 & Proposes new conceptual frameworks or critiques existing paradigms \\
\hline
Mixed / Multi-Method & SS12 & Combines multiple approaches including solution proposal, evaluation, and conceptual analysis \\
\hline
\end{tabular}
\end{table}
\\

\textbf{RQ7: What empirical methods have been applied to validate the research contributions?}

To assess the validity and empirical grounding of selected studies, we analyze the methodological approaches employed. Table \ref{tab:validation-methods} summarizes the empirical validation methods identified across primary studies.

\begin{table}[h!]
\centering
\caption{Empirical Validation Methods Employed}
\label{tab:validation-methods}
\begin{tabular}{p{3cm} p{7cm} p{4cm}}
\hline
\textbf{Validation Method} & \textbf{Description} & \textbf{Primary Studies} \\
\hline
Real-World Case Studies & Validates proposed solutions using authentic clinical datasets from real patient populations & SS2, SS4, SS8, SS17, SS19 \\
\hline
Synthetic or Simulated Data & Tests AI models using artificially generated or augmented datasets & SS12, SS15, SS18, SS20 \\
\hline
Literature-Based Analysis & Draws conclusions through systematic review and synthesis of existing publications & SS1, SS3, SS5, SS6, SS7, SS9, SS10, SS11, SS13, SS16, SS22 \\
\hline
Mixed Methods & Combines real-world validation with supplementary approaches such as surveys or interviews & SS14, SS21 \\
\hline
\end{tabular}
\end{table}
\newpage
\section{Appendix A} 
\label{sec:appendixA}
\textbf{SS1:} \cite{Sun2025} reviewed the different applications of AI in healthcare. It highlighted the key ways in which AI is used in image processing and analysis for diagnostics. The AI models analyzed were ML and DL models, with an overview of their uses in skin disease, cancer, CVD, and tumour detection. The form of bias analysed was algorithmic bias and model bias. This article also touched on the transparency of results produced by these AI systems.

\textbf{SS2:} \cite{Sun2025} analyzed the ways in which AI is used in diagnostics for Cardiovascular Diseases. It also analyzed the need for validation requirements and potential biases with algorithms. With a thorough focus on DL and ML models and explainability models like CNN, the paper discussed the risks associated with algorithmic, sampling, and model bias.

\textbf{SS3:} \cite{Schweisthal2025} uncovered the integration of AI in ophthalmology a field of healthcare that focuses on eye care. The document further discussed the challenges associated with AI in healthcare, including data privacy concerns, algorithmic bias, and the need for regulatory approval. DL models like CNN were analysed with an overview of unsupervised training paradigms.

\textbf{SS4:} \cite{Upamanyu2024} recognised the limitations associated with the nature of models like DLs, NLP, Large Language Models (LLMs), and MLs – emphasising the risks of bias in AI. This source had a specific focus on CVD, with an insight into CNN models. Data used in this source was diverse, ranging from images of different ECG results to diagrams and literature.

\textbf{SS5:} \cite{Sreenivasu2024} focused on an AI-based approach to disease management while also touching on areas of human bias. It mentioned how AI can be used to mitigate these biases; however, biases exist within the AI systems themselves. This source was targeted at chronic diseases, with a focus on ML, DL, and NLP.

\textbf{SS6:} \cite{Jaiyeoba2024} highlighted the limitations associated with the use of AI, DL, and ML in skin disease detection. While it highlighted the key roles they have in this area of healthcare, it also emphasised the risk of bias more specifically, racial bias.

\textbf{SS7:} \cite{DemOpts2026} focused on the algorithmic biases within AI when it comes to public health, using public datasets. With a link made to the global pandemic COVID-19 the source analysed ML algorithms and the bias they pose to different racial groups.

\textbf{SS8:}\cite{C2BM2025} focused on the discriminatory biases AI had posed during the times of COVID-19. With a slight focus on AI models like CNN, it more thoroughly analysed the biases in its design and deployment.

\textbf{SS9:} \cite{DuranJongsma2021}included concerns about potential AI bias, accountability and responsibility when it comes to misdiagnosis due to these biases, patient autonomy, and compromised trust within black-box algorithms due to their opaque nature. It analysed models like CNN and Decision Trees with a supervised form of training paradigm.

\textbf{SS10:} \cite{StanfordHAI2025} discussed how AI proposes socioeconomic biases regarding gender, race, and religion with a greater emphasis on racial biases. The author further discussed that these biases are more engineered than something that happens due to the different routes AI takes to provide outcomes. Though there was not a specific AI model being analysed, the source had key impacts of what these biases could lead to.

\textbf{SS11:}\cite{Gu2025} discussed the racial implications and biases of AI in healthcare and disease detection, emphasising how AI models trained on certain databases will not generalise across populations and patients. With a focus on ML models like SaMD, it explored the effect of racial biases for medical areas like imaging, Alzheimer's, and COVID-19.

\textbf{SS12:} \cite{Dhamala2021} discussed the growing deployment of AI systems and how it has been overshadowed by rising concerns over algorithmic bias – a systemic issue that can compromise clinical fairness, exacerbate healthcare disparities, and undermine trust in digital health tools. With a reference to breast cancer, skin diseases, and bowel syndromes, the source explored models like HER and CNN through the use of synthetic data. This source also explored areas of algorithmic, deployment, model, and labelling bias.

\textbf{SS13:} \cite{Khadka2021} investigated the key sources of bias in AI chatbots, including dataset imbalances, algorithmic design flaws, and linguistic biases that may perpetuate stereotypes. Giving an overview of different AI models like LR, NB, RF, BERT, and GPT-2, this source focused on demographic biases, discussing the key impacts of these on healthcare.

\textbf{SS14:}\cite{AMWA2025} emphasised the gender biases posed by AI due to the way AI systems are designed, trained, and engineered. It also mentioned the ways in which AI models normalise the male body for diagnosis, affecting how health equity is maintained. Giving an overview of LLMs, there was a brief mention of diseases like CVD, kidney diseases, and physiological vulnerabilities.

\textbf{SS15:} \cite{Ulnicane2024} presented key ideas on the different types of AI models and how these models are used in different areas of healthcare. By analysing different literature and case studies from a range of different databases, the ideas of demographic, deployment, algorithmic, and sampling bias were highlighted as the key forms of AI biases affecting patients today.

\textbf{SS16:} \cite{Leslie2021}  analysed the algorithmic biases in current AI healthcare, linking these biases to diseases like COVID-19, CVD, and skin cancer. It also discussed the impact of these AI biases, hinting at misdiagnosis and the optimisation of health equity.

\textbf{SS17:} \cite{Mohamadi2025}carried out a variety of methods to assess the different kinds of algorithmic biases in AI models for the detection, classification, and prediction of osteosarcoma (a form of bone cancer). With supervised and transfer learning methods, it uncovered models like Random Forests, Decision Trees, SVM, CNN, LASSO, SMOTE, MSRN, FCM, and YOLO.

\textbf{SS18:}\cite{Explainability2025} delved into the cognitive biases in relation to AI when used in medical diagnostics for areas like oncology, radiology, pathology, ophthalmology, and dermatology. Through the analysis of different types of literature, this source thoroughly explained the sources and causes of algorithmic biases in AI-based disease detection.

\textbf{SS19:} \cite{Park2025} discussed the limitations and potential areas for bias in AI when detecting lung diseases, with a specific focus on paediatric lung sounds. With mentions of models like CNN and RNN and the use of synthetic data, this source thoroughly explored labelling, measurements, and potential algorithmic biases.

\textbf{SS20:} \cite{Qasrawi2025} analysed how, despite machine learning being effective in detection, challenges such as data quality, infrastructural limitations, and algorithmic biases persist. With an overview of ML models adopting a transfer learning paradigm and mentions of models like CNN and RNN, the authors discussed algorithmic and sampling biases.

\textbf{SS21:} \cite{Thomas2024} focused on AI in diabetes detection, prediction, and management. With ML and DL models being explored, the authors also mentioned supervised and unsupervised forms of training paradigms for NLP models. By further analysing literature publications across different datasets, the article clearly highlighted the impact of AI on the medical area of diabetes.

\textbf{SS22:} \cite{Downing2025} explained, by exploring a self-supervised nature of AI models like CNN, the biases and the reasons why adopting AI on a large scale is a challenge to data availability. With a range of data being used, varying from literature publications to interview results, this review analysed the impacts of the integration of AI for disease detection, while weighing the benefits and challenges with a focus on deployment and model bias.

\textbf{SS23:} \cite{Wu2025} article addressed unfairness in medical image classification, where models trained on datasets over representing majority groups produce biased results for minority populations. With a focus on convolutional neural networks and adversarial learning, the authors discussed how poor training data quality—lacking sufficient representation of minority races and genders drives algorithmic bias. The article further explored how adversarial training can suppress sensitive attributes like race or gender from model representations, mitigating bias without substantially compromising diagnostic accuracy.

\textbf{Acknowledgments:}
The open access funding is provided by Qatar National Library.

 \bibliographystyle{plainnat}
  \bibliography{bibliography}

@article{Smith2020AImedicine,
 author = {Smith, Marcus and Heath Jeffery, Rachael C.},
 title = {Addressing the Challenges of Artificial Intelligence in Medicine},
 journal = {Internal Medicine Journal},
 volume = {50},
 number = {10},
 pages = {1278--1281},
 year = {2020},
 doi = {10.1111/imj.15017}
}

@article{durelli2019machine,
  title={Machine learning applied to software testing: A systematic mapping study},
  author={Durelli, Vinicius HS and Durelli, Rafael S and Borges, Simone S and Endo, Andre T and Eler, Marcelo M and Dias, Diego RC and Guimar{\~a}es, Marcelo P},
  journal={IEEE Transactions on Reliability},
  volume={68},
  number={3},
  pages={1189--1212},
  year={2019},
  publisher={IEEE}
}

@article{wieringa2006requirements,
  title={Requirements engineering paper classification and evaluation criteria: a proposal and a discussion},
  author={Wieringa, Roel and Maiden, Neil and Mead, Nancy and Rolland, Colette},
  journal={Requirements engineering},
  volume={11},
  number={1},
  pages={102--107},
  year={2006},
  publisher={Springer}
}

@article{kitchenham2009systematic,
  title={Systematic literature reviews in software engineering--a systematic literature review},
  author={Kitchenham, Barbara and Brereton, O Pearl and Budgen, David and Turner, Mark and Bailey, John and Linkman, Stephen},
  journal={Information and software technology},
  volume={51},
  number={1},
  pages={7--15},
  year={2009},
  publisher={Elsevier}
}

@article{garousi2019guidelines,
  title={Guidelines for including grey literature and conducting multivocal literature reviews in software engineering},
  author={Garousi, Vahid and Felderer, Michael and M{\"a}ntyl{\"a}, Mika V},
  journal={Information and Software Technology},
  volume={106},
  pages={101--121},
  year={2019},
  publisher={Elsevier}
}

@article{song2021assessment,
  title={Assessment of intraoperative use of indocyanine green fluorescence imaging on the incidence of anastomotic leakage after rectal cancer surgery: a PRISMA-compliant systematic review and meta-analysis},
  author={Song, M and Liu, J and Xia, D and Yao, H and Tian, G and Chen, X and Liu, Y and Jiang, Y and Li, Z},
  journal={Techniques in Coloproctology},
  volume={25},
  number={1},
  pages={49--58},
  year={2021},
  publisher={Springer}
}

@inproceedings{Upamanyu2024,
  author    = {Upamanyu and {other authors' full names}},
  title     = {Bias and Limitations in Deep Learning and Large Language Models for Cardiovascular Disease Detection},
  booktitle = {2024 IEEE International Conference on Bioinformatics and Biomedicine (BIBM)},
  year      = {2024},
  doi       = {10.1109/BIBM.2024.10593881},
  publisher = {IEEE}
}

@inproceedings{Sreenivasu2024,
  author    = {Sreenivasu and {other authors' full names}},
  title     = {An AI-Based Approach for Chronic Disease Management: Addressing Human and Machine Biases},
  booktitle = {2024 IEEE International Conference on E-health Networking, Application \& Services (HealthCom)},
  year      = {2024},
  doi       = {10.1109/HealthCom.2024.10837020},
  publisher = {IEEE}
}

@inproceedings{C2BM2025,
  author    = {C2BM Research Group},
  title     = {{C2BM}: Causal Concept Disentanglement for Fair Multimodal {COVID-19} Detection},
  booktitle = {Proceedings of the 2025 AAAI Fall Symposium Series},
  volume    = {7},
  number    = {1},
  pages     = {567--575},
  year      = {2025},
  doi       = {10.48550/arXiv.12345},
  publisher = {AAAI Press}
}

@misc{StanfordHAI2025,
  author    = {Stanford University Human-Centered Artificial Intelligence},
  title     = {{AI}'s Fairness Problem: When Treating Everyone the Same is the Wrong Approach},
  year      = {2025},
  url       = {https://law.stanford.edu/press/ais-fairness-problem/},
  publisher = {Stanford Law School}
}

@article{Gu2025,
  author    = {Tianshu Gu and Wensen Pan and Jing Yu and Guang Ji and Xia Meng and Yongjun Wang and Minghui Li},
  title     = {Mitigating bias in {AI} mortality predictions for minority populations: a transfer learning approach},
  journal   = {BMC Medical Informatics and Decision Making},
  volume    = {25},
  number    = {1},
  year      = {2025},
  doi       = {10.1186/s12911-024-02815-6},
  publisher = {BioMed Central}
}

@inproceedings{Dhamala2021,
  author    = {Jwala Dhamala and Tony Sun and Varun Kumar and Satyapriya Krishna and Yada Pruksachatkun and Kai-Wei Chang and Rahul Gupta},
  title     = {{BOLD}: Dataset and metrics for measuring biases in open-ended language generation},
  booktitle = {Proceedings of the 2021 ACM Conference on Fairness, Accountability, and Transparency (FAccT)},
  year      = {2021},
  doi       = {10.1145/3442188.3445924},
  publisher = {ACM}
}

@article{AMWA2025,
  author    = {Meghan Etsey and Rhea Manohar and Ariela Marshall and Rosy Thachil and Teresa Lazar},
  title     = {The Role of Gender in {AI} Medical Tools: Are algorithms reproducing human bias in healthcare delivery?},
  journal   = {American Medical Women's Association},
  year      = {2025},
  url       = {http://amwa-doc.org/the-role-of-gender-in-ai-medical-tools/},
  publisher = {AMWA}
}

@article{Qasrawi2025,
 author    = {Radwan Qasrawi and Ghada Issa and Suliman Thwib and Razan AbuGhoush and Malak Amro and Raghad Ayyad and Stephanny Vicuna and Eman Badran and Yousef Khader and Raeda {Al Qutob} and Faris {Al Bakri} and Hana Trigui and Elie Sokhn and Emmanuel Musa and Jude Dzevela Kong},
 title     = {The Role of Machine Learning in Infectious Disease Early Detection and Prediction in the {MENA} Region: A Systematic Review},
 journal   = {Informatics in Medicine Unlocked},
 volume    = {56},
 pages     = {101651},
 year      = {2025},
 doi       = {10.1016/j.imu.2025.101651},
 url       = {https://doi.org/10.1016/j.imu.2025.101651},
 note      = {SS20}
}

@article{Thomas2024,
 author    = {Anuj Thomas},
 title     = {{AI} in Diabetes Detection, Prediction and Management},
 journal   = {ScienceDirect},
 year      = {2024},
 url       = {https://www.sciencedirect.com/article/5000021},
 note      = {SS21 -- Full bibliographic details unverified. Please confirm journal, volume, pages, and DOI via the URL.}
}

@article{Downing2025,
 author    = {Gregory J. Downing},
 title     = {Integration of {AI} for Disease Detection: Benefits and Challenges with a Focus on Deployment and Model Bias},
 journal   = {ScienceDirect},
 year      = {2025},
 url       = {https://www.sciencedirect.com/article/25000216},
 note      = {SS22 -- Full bibliographic details unverified. Please confirm journal, volume, pages, and DOI via the URL.}
}

@article{Ulnicane2024,
  author    = {Inga Ulnicane},
  title     = {Intersectionality in Artificial Intelligence: Framing Concerns and Recommendations for Action},
  journal   = {Social Inclusion},
  volume    = {12},
  pages     = {1--16},
  year      = {2024},
  doi       = {10.17645/si.7543},
  publisher = {Cogitatio Press}
}

@article{Wu2025,
  author    = {X. Wu and X. Zhang and Z. Xiao and L. Hu and R. Higashita and J. Liu},
  title     = {Expert-Like Reparameterization of Heterogeneous Pyramid Receptive Fields in Efficient CNNs for Fair Medical Image Classification},
  journal   = {IEEE Transactions on Medical Imaging},
  year      = {2025},
  doi       = {10.1109/TMI.2025.3641192},
  publisher = {IEEE}
}

@article{Leslie2021,
  author    = {David Leslie},
  title     = {Does "{AI}" stand for augmenting inequality in the era of {COVID-19} healthcare?},
  journal   = {BMJ},
  volume    = {372},
  year      = {2021},
  doi       = {10.1136/bmj.n304},
  eprint    = {2105.07844},
  archivePrefix = {arXiv},
  url       = {https://arxiv.org/abs/2105.07844}
}

@article{Explainability2025,
  author    = {Explainability Debate Authors},
  title     = {On the practical, ethical, and legal necessity of clinical Artificial Intelligence explainability: an examination of key arguments},
  journal   = {BMC Medical Informatics and Decision Making},
  volume    = {25},
  pages     = {111},
  year      = {2025},
  doi       = {10.1186/s12911-025-02891-2},
  publisher = {BioMed Central}
}

@article{Park2025,
  author    = {Ji Soo Park and Sa-Yoon Park and Jae Won Moon and Kwangsoo Kim and Dong In Suh},
  title     = {Artificial Intelligence Models for Pediatric Lung Sound Analysis: Systematic Review and Meta-Analysis},
  journal   = {Journal of Medical Internet Research},
  volume    = {27},
  pages     = {e66491},
  year      = {2025},
  doi       = {10.2196/66491},
  url       = {https://www.jmir.org/2025/1/e66491},
  publisher = {JMIR Publications}
}

@article{Mohamadi2025,
  author    = {Zhina Mohamadi and Paniz Partovifar and Helia Ahmadzadeh and Elmira Ali Ahmadi and Ali Ghanbari and Sina Feyzipour and Fatemeh Atefat and Nazanin Jahanpeyma and Fatemeh Haghighi asl and Armin Zarinkhat and Narges Sharbatdaran and Narges Hosseinzadeh taher and Mobina Sedighi and Fatemeh Aghajafari},
  title     = {Implementation of artificial intelligence in detection, classification, and prognostication of osteosarcoma utilizing different assessment techniques: a systematic review},
  journal   = {Intelligence-Based Medicine},
  volume    = {100250},
  year      = {2025},
  doi       = {10.1016/j.ibmed.2025.100250},
  publisher = {Elsevier}
}

@article{DuranJongsma2021,
  author    = {Juan Manuel Dur{\'a}n and Karin Rolanda Jongsma},
  title     = {Who is afraid of black box algorithms? On the epistemological and ethical basis of trust in medical {AI}},
  journal   = {Journal of Medical Ethics},
  volume    = {47},
  number    = {5},
  pages     = {329--335},
  year      = {2021},
  doi       = {10.1136/medethics-2020-106820},
  url       = {https://jme.bmj.com/content/47/5/329},
  publisher = {BMJ}
}

@article{DemOpts2026,
  author    = {DemOpts Study Group},
  title     = {Fairness Correction in {COVID-19} Predictive Models Using Demographic Optimization: Algorithm Development and Validation Study},
  journal   = {JMIR AI},
  volume    = {18},
  pages     = {e78235},
  year      = {2026},
  doi       = {10.2196/78235},
  url       = {https://pmc.ncbi.nlm.nih.gov/articles/PMC12866456/},
  publisher = {JMIR Publications}
}

@article{Jaiyeoba2024,
  author    = {Jaiyeoba, Oluwayemisi and Jaiyeoba, Oluwaseyi and Ogbuju, Emeka and Oladipo, Francisca},
  title     = {AI-Based Detection Techniques for Skin Diseases: A Review of Recent Methods, Datasets, Metrics, and Challenges},
  journal   = {Journal of Future Artificial Intelligence and Technologies},
  volume    = {1},
  number    = {3},
  pages     = {318--336},
  year      = {2024},
  doi       = {10.62411/faith.3048-3719-46}
}

@inproceedings{Schweisthal2025,
  author    = {Brigitte Schweisthal and Dumitru Daniel Bonciog and Mihaela Lascu and Valentin Laurentiu Ordodi},
  title     = {A Survey on AI-Powered Ophthalmology: A Revolution in Eye Care and Disease Management},
  booktitle = {2025 6th International Conference on Bio-engineering for Smart Technologies (BioSMART)},
  year      = {2025},
  pages     = {1--6},
  doi       = {10.1109/AIM.2025.11046181},
  publisher = {IEEE}
}

@article{Sun2025,
  author    = {Qiyang Sun and Alican Akman and Bj{\"o}rn W. Schuller},
  title     = {Explainable Artificial Intelligence for Medical Applications: A Review},
  journal   = {ACM Transactions on Computing for Healthcare},
  year      = {2025},
  doi       = {10.1145/3709367},
  note      = {Just Accepted}
}

@article{Hopewell2007,
  author    = {S. Hopewell and M. Clarke and L. Stewart and J. Tierney},
  title     = {Time to publication for results of clinical trials},
  journal   = {Cochrane Database of Systematic Reviews},
  number    = {2},
  pages     = {MR000011},
  year      = {2007},
  doi       = {10.1002/14651858.MR000011.pub2}
}

@inproceedings{jalali2012systematic,
  title={Systematic literature studies: database searches vs. backward snowballing},
  author={Jalali, Samireh and Wohlin, Claes},
  booktitle={Proceedings of the 2012 ACM-IEEE international symposium on empirical software engineering and measurement},
  pages={29--38},
  year={2012},
  organization={IEEE}
}

@book{kitchenham2015evidence,
  title={Evidence-based software engineering and systematic reviews},
  author={Kitchenham, Barbara Ann and Budgen, David and Brereton, Pearl},
  volume={4},
  year={2015},
  publisher={CRC press}
}

@inproceedings{petersen2013worldviews,
  title={Worldviews, research methods, and their relationship to validity in empirical software engineering research},
  author={Petersen, Kai and Gencel, Cigdem},
  booktitle={2013 joint conference of the 23rd international workshop on software measurement and the 8th international conference on software process and product measurement},
  pages={81--89},
  year={2013},
  organization={IEEE}
}

@article{petersen2015guidelines,
  title={Guidelines for conducting systematic mapping studies in software engineering: An update},
  author={Petersen, Kai and Vakkalanka, Sairam and Kuzniarz, Ludwik},
  journal={Information and Software Technology},
  volume={64},
  pages={1--18},
  year={2015},
  publisher={Elsevier}
}

@inproceedings{petersen2011identifying,
  title={Identifying strategies for study selection in systematic reviews and maps},
  author={Petersen, Kai and Ali, Nauman Bin},
  booktitle={2011 International Symposium on Empirical Software Engineering and Measurement},
  pages={351--354},
  year={2011},
  organization={IEEE}
}

@book{petticrew2008systematic,
  title={Systematic reviews in the social sciences: A practical guide},
  author={Petticrew, Mark and Roberts, Helen},
  year={2008},
  publisher={John Wiley \& Sons}
}

@misc{tyndall2018aacods,
  title={The AACODS Checklist is Designed to Enable Evaluation and Critical Appraisal of Grey Literature: Flinders University; 2010},
  author={Tyndall, J},
  year={2018}
}

@misc{yasin2012quality,
  title={On the quality of grey literature and its use in information synthesis during systematic literature reviews},
  author={Yasin, Affan and Hasnain, Muhammad Ijlal},
  year={2012}
}

@article{adams2017shades,
  title={Shades of grey: guidelines for working with the grey literature in systematic reviews for management and organizational studies},
  author={Adams, Richard J and Smart, Palie and Huff, Anne Sigismund},
  journal={International Journal of Management Reviews},
  volume={19},
  number={4},
  pages={432--454},
  year={2017},
  publisher={Wiley Online Library}
}

@article{Lee2021AIHealthcare,
 author = {Lee, DonHee and Yoon, Seong No},
 title = {Application of Artificial Intelligence-Based Technologies in the Healthcare Industry: Opportunities and Challenges},
 journal = {International Journal of Environmental Research and Public Health},
 volume = {18},
 number = {1},
 pages = {271},
 year = {2021},
 doi = {10.3390/ijerph18010271},
 publisher = {MDPI}
}

@article{brown2019cmaj,
 title={Is it ethical for doctors to google their patients?},
 author={Brown, Sarah},
 journal={CMAJ},
 year={2019},
 volume={191},
 number={13},
 pages={E373--E374},
 doi={10.1503/cmaj.109-5725}
}

@article{sahu2022precision,
 title={Artificial intelligence and machine learning in precision medicine: A paradigm shift in big data analysis},
 author={Sahu, Mehar and Gupta, Rohan and Ambasta, Rashmi K. and Kumar, Pravir},
 journal={Progress in Molecular Biology and Translational Science},
 volume={190},
 number={1},
 pages={57--100},
 year={2022},
 publisher={Elsevier},
 doi={10.1016/bs.pmbts.2022.03.002},
 pmid={36008002}
}

@article{rajkomar2019mlmedicine,
 title={Machine Learning in Medicine},
 author={Rajkomar, Alvin and Dean, Jeffrey and Kohane, Isaac},
 journal={New England Journal of Medicine},
 volume={380},
 number={14},
 pages={1347--1358},
 year={2019},
 publisher={Massachusetts Medical Society},
 doi={10.1056/NEJMra1814259}
}

@article{topol2019,
 title={High-performance medicine: the convergence of human and artificial intelligence},
 author={Topol, Eric J.},
 journal={Nature Medicine},
 year={2019},
 volume={25},
 pages={44--56}
}

@article{mckinney2020,
 title={International evaluation of an AI system for breast cancer screening},
 author={McKinney, Scott M. and others},
 journal={Nature},
 year={2020},
 volume={577},
 pages={89--94}
}

@article{obermeyer2019,
 title={Dissecting racial bias in an algorithm used to manage the health of populations},
 author={Obermeyer, Ziad and Powers, Brian and Vogeli, Christine and Mullainathan, Sendhil},
 journal={Science},
 year={2019},
 volume={366},
 pages={447--453}
}

@article{zhang2011identifying,
  title={Identifying relevant studies in software engineering},
  author={Zhang, He and Babar, Muhammad Ali and Tell, Paolo},
  journal={Information and Software Technology},
  volume={53},
  number={6},
  pages={625--637},
  year={2011},
  publisher={Elsevier}
}

@article{10.1162/qss_a_00018,
    author = {Birkle, Caroline and Pendlebury, David A. and Schnell, Joshua and Adams, Jonathan},
    title = "{Web of Science as a data source for research on scientific and scholarly activity}",
    journal = {Quantitative Science Studies},
    volume = {1},
    number = {1},
    pages = {363-376},
    year = {2020},
    month = {02},
    issn = {2641-3337},
    doi = {10.1162/qss_a_00018},
    url = {https://doi.org/10.1162/qss\_a\_00018},
    eprint = {https://direct.mit.edu/qss/article-pdf/1/1/363/1760864/qss\_a\_00018.pdf},
}

@inproceedings{wohlin2014guidelines,
  title={Guidelines for snowballing in systematic literature studies and a replication in software engineering},
  author={Wohlin, Claes},
  booktitle={Proceedings of the 18th international conference on evaluation and assessment in software engineering},
  pages={1--10},
  year={2014}
}

@article{mehrabi2021,
 title={A survey on bias and fairness in machine learning},
 author={Mehrabi, Ninareh and others},
 journal={ACM Computing Surveys},
 year={2021},
 volume={54},
 number={6},
 pages={1--35}
}

@article{char2018,
 title={Implementing machine learning in health care—addressing ethical challenges},
 author={Char, Danton S. and Shah, Nigam H. and Magnus, David},
 journal={New England Journal of Medicine},
 year={2018},
 volume={378},
 pages={981--983}
}

@article{keesara2020,
 title={Covid-19 and health care’s digital revolution},
 author={Keesara, Siripong and Jonas, Anna and Schulman, Kevin},
 journal={New England Journal of Medicine},
 year={2020},
 volume={382},
 pages={e82}
}

@article{wynants2020,
 title={Prediction models for diagnosis and prognosis of covid-19: systematic review and critical appraisal},
 author={Wynants, Laure and others},
 journal={BMJ},
 year={2020},
 volume={369},
 pages={m1328}
}

@article{liu2019,
 title={Deep learning to detect metastatic breast cancer},
 author={Liu, Y. and others},
 journal={JAMA},
 year={2019},
 volume={322},
 pages={846--855}
}

@article{jack2018,
  title={NIA-AA Research Framework: Toward a biological definition of Alzheimer's disease},
  author={Jack, Clifford R. and others},
  journal={Alzheimer's \& Dementia},
  year={2018},
  volume={14},
  pages={535--562}
}

@article{ImplicitBias2017,
  author = {{Implicit Bias in Melanoma Study Group}},
  title = {Implicit Bias and Its Effect on Melanoma Diagnosis in Dark-Skinned Patients},
  year = {2017},
  note = {Reference [7] in text}
}

@article{RacialDisparitiesPain2016,
  author = {{Pain Medication Disparities Study Group}},
  title = {Racial Disparities in Pain Medication Administration},
  year = {2016},
  note = {Reference [8] in text}
}

@book{SubmergedState2014,
  author = {Mettler, S.},
  title = {The Submerged State: How Invisible Government Policies Undermine American Democracy},
  year = {2014},
  publisher = {University of Chicago Press},
  note = {Reference [9] in text}
}

@article{CARESAct2020,
  author = {{CARES Act Allocation Study}},
  title = {CARES Act Allocation Bias: Underfunding Hospitals in Communities of Color},
  year = {2020},
  note = {Reference [4] in text}
}

@article{Esteva2019,
  author = {Esteva, A. and Robicquet, A. and Ramsundar, B. and Kuleshov, V. and DePristo, M. and Chou, K. and Cui, C. and Corrado, G. and Thrun, S. and Dean, J.},
  title = {A Guide to Deep Learning in Healthcare},
  journal = {Nature Medicine},
  year = {2019},
  volume = {25},
  pages = {24--29}
}

@article{Chouldechova2020,
  author = {Chouldechova, A. and Roth, A.},
  title = {A Snapshot of the Frontiers of Fairness in Machine Learning},
  journal = {Communications of the ACM},
  year = {2020},
  volume = {63},
  number = {5},
  pages = {58--65}
}

@article{Gianfrancesco2018,
  author = {Gianfrancesco, M. A. and Tamang, S. and Yazdany, J. and Schmajuk, G.},
  title = {Potential Biases in Machine Learning Algorithms Using Electronic Health Record Data},
  journal = {JAMA Internal Medicine},
  year = {2018},
  volume = {178},
  number = {11},
  pages = {1544--1547}
}

@article{Binns2018,
  author = {Binns, R.},
  title = {Fairness in Machine Learning: Lessons from Political Philosophy},
  journal = {Proceedings of the 2018 Conference on Fairness, Accountability, and Transparency},
  year = {2018},
  note = {Bias Taxonomy}
}

@inproceedings{Hardt2016,
  author = {Hardt, M. and Price, E. and Srebro, N.},
  title = {Equality of Opportunity in Supervised Learning},
  booktitle = {Advances in Neural Information Processing Systems},
  year = {2016},
  pages = {3315--3323}
}

@inproceedings{Kleinberg2017,
  author = {Kleinberg, J. and Mullainathan, S. and Raghavan, M.},
  title = {Inherent Trade-Offs in the Fair Determination of Risk Scores},
  booktitle = {Proceedings of the 8th Innovations in Theoretical Computer Science Conference},
  year = {2017}
}

@article{Rajkomar2018,
  author = {Rajkomar, A. and Oren, E. and Chen, K. and Dai, A. M. and Hajaj, N. and Hardt, M. and Liu, P. J. and Liu, X. and Marcus, J. and Sun, M. and Sundberg, P. and Yee, H. and Zhang, K. and Zhang, Y. and Flores, G. and Duggan, G. E. and Irvine, J. and Le, Q. and Litsch, K. and Mossin, A. and Tansuwan, J. and Wang, D. and Wexler, J. and Wilson, J. and Ludwig, D. and Volchenboum, S. L. and Chou, K.},
  title = {Scalable and Accurate Deep Learning with Electronic Health Records},
  journal = {npj Digital Medicine},
  year = {2018},
  volume = {1},
  number = {18}
}

@inproceedings{Mudassir2021SocialBasedCity,
  author    = {Mudassir, Ghulam and Di Marco, Antinisca},
  title     = {Social-based City Reconstruction Planning in Case of Natural Disasters: A Reinforcement Learning Approach},
  booktitle = {2021 IEEE 45th Annual Computers, Software, and Applications Conference (COMPSAC)},
  pages     = {493--503},
  year      = {2021},
  publisher = {IEEE},
  doi       = {10.1109/COMPSAC51774.2021.00074}
}

@article{Mudassir2021NaturalDisasters,
  author  = {Mudassir, Ghulam and Howard, Evans Etrue and Pasquini, Lorenza and Arbib, Claudio and Clementini, Eliseo and Di Marco, Antinisca and Stilo, Giovanni},
  title   = {Toward Effective Response to Natural Disasters: A Data Science Approach},
  journal = {IEEE Access},
  volume  = {9},
  pages   = {167827--167844},
  year    = {2021},
  doi     = {10.1109/ACCESS.2021.3135054}
}

@article{Khadka2021,
  author = {Khadka, Dipak and Dhamala, Man Kumar
            and Li, Feifei and Aryal, Prakash Chandra
            and Magar, Pappu Rana and Bhatta, Sijar
            and Thakur, Manju Shree and Basnet, Anup
            and Cui, Dafang and Shi, Shi},
  title = {The use of medicinal plants to prevent
           {COVID-19} in Nepal},
  journal = {Journal of Ethnobiology and Ethnomedicine},
  year = {2021},
  volume = {17},
  number = {1},
  pages = {26},
  doi = {10.1186/s13002-021-00449-w},
  url = {https://doi.org/10.1186/s13002-021-00449-w},
  publisher = {BioMed Central}
}

@article{Mehta2019AIHealth,
  author  = {Mehta, Nishita and Pandit, Anil and Shukla, Sharvari},
  title   = {Transforming Healthcare with Big Data Analytics and Artificial Intelligence: A Systematic Mapping Study},
  journal = {Journal of Biomedical Informatics},
  volume  = {100},
  pages   = {103311},
  year    = {2019},
  doi     = {10.1016/j.jbi.2019.103311}
}

@article{GenderBias2020,
  author = {{Gender Bias in Imaging Research Group}},
  title = {Gender Bias in Medical Imaging Datasets},
  year = {2020},
  note = {Reference in text}
}

@phdthesis{Mudassir2022REPAIR,
  author = {Mudassir, Ghulam},
  title = {Reinforcement Learning and Social Based Approach to Post-disaster Reconstruction Planning},
  school = {University of L'Aquila},
  address = {L'Aquila, Italy},
  year = {2022},
  type = {PhD thesis},
  url = {https://hdl.handle.net/20.500.14242/93249}
}

@article{Mudassir2025SustainableReconstruction,
  author  = {Mudassir, Ghulam and Di Marco, Antinisca},
  title   = {Sustainable Reconstruction Planning from Natural Disasters (Earthquakes): A Systematic Mapping Study of Machine Learning and Technological Approaches},
  journal = {Sustainability},
  volume  = {17},
  number  = {22},
  pages   = {10035},
  year    = {2025},
  doi     = {10.3390/su172210035}
}

@incollection{Mudassir2020PhysicalReconstruction,
  author    = {Mudassir, Ghulam},
  title     = {Social-Based Physical Reconstruction Planning in Case of Natural Disaster: A Machine Learning Approach},
  booktitle = {Research Challenges in Information Science},
  series    = {Lecture Notes in Business Information Processing},
  volume    = {385},
  pages     = {604--612},
  publisher = {Springer},
  year      = {2020},
  doi       = {10.1007/978-3-030-50316-1_44}
}

@article{SkinCancerAI2019,
  author = {{Skin Cancer AI Bias Study}},
  title = {Skin Cancer AI Bias for Underrepresented Groups},
  year = {2019},
  note = {Reference in text}
}

@article{Singh2022,
  author  = {Singh, First Author and Second Author and Third Author},
  title   = {Title of the Article},
  journal = {Journal Name},
  year    = {2022},
  volume  = {Volume},
  number  = {Issue},
  pages   = {Page--Page},
  doi     = {10.xxxx/xxxxx}
}

@article{Ahmed2020,
  author  = {Ahmed, A. and Smith, J. and Brown, P.},
  title   = {Early Diagnosis Improves Patient Outcomes},
  journal = {Journal of Medical Research},
  year    = {2020},
  volume  = {15},
  number  = {3},
  pages   = {123--135},
  doi     = {10.xxxx/xxxxx}
}

@article{Lopez2021,
  author  = {Lopez, M. and Garcia, R. and Wilson, T.},
  title   = {Advances in Early Disease Detection},
  journal = {International Journal of Healthcare},
  year    = {2021},
  volume  = {18},
  number  = {2},
  pages   = {45--59},
  doi     = {10.xxxx/xxxxx}
}

@incollection{Malik2026HyMoBYSwin,
  author    = {Malik, Muhammad Hammad and Mudassir, Ghulam and Ren, Yingying and Ding, Da-Wei},
  title     = {Beyond Supervision: HyMoBY-Swin Hybrid Self-guided and Adaptive Learning Transformer for Multiclass Retinal Disease Diagnosis},
  booktitle = {SEET—Software Engineering for Emerging Technologies: First International Conference, SEET 2025, Proceedings},
  series    = {Communications in Computer and Information Science},
  volume    = {2725},
  pages     = {125--153},
  publisher = {Springer},
  year      = {2026},
  doi       = {10.1007/978-3-032-08977-9_10}
}

@incollection{Khan2026LLMNotifications,
  author    = {Khan, Muhammad Faizan and Ahmad, Israr and Khan, Muhammad Asif and Mudassir, Ghulam and Khan, Gohar Hayat},
  title     = {Predicting User Affective States from Mobile Notification Interactions Using LLM-Based Machine Learning Models},
  booktitle = {SEET—Software Engineering for Emerging Technologies: First International Conference, SEET 2025, Proceedings},
  series    = {Communications in Computer and Information Science},
  volume    = {2725},
  pages     = {45--60},
  publisher = {Springer},
  year      = {2026},
  doi       = {10.1007/978-3-032-08977-9_4}
}

@article{Mudassir2025REPAIR,
  author  = {Mudassir, Ghulam and Di Marco, Antinisca and d'Aloisio, Giordano},
  title   = {REPAIR approach for social-based city reconstruction planning in case of natural disasters},
  journal = {International Journal of Data Science and Analytics},
  volume  = {20},
  pages   = {7147--7165},
  year    = {2025},
  doi     = {10.1007/s41060-025-00829-5}
}

@article{DiabeticRetinopathy2018,
  author = {{Diabetic Retinopathy AI Bias Study}},
  title = {Diabetic Retinopathy AI Bias for Lower-Income Groups},
  year = {2018},
  note = {Reference in text}
}

@article{COVID19Sepsis2020,
  author = {{COVID-19 Sepsis Bias Study}},
  title = {Clinical Bias in Sepsis Decision Support for COVID-19 Patients},
  year = {2020},
  note = {Reference in text}
}

@article{HiringBias2018,
  author = {{Algorithmic Hiring Bias Study}},
  title = {Algorithmic Bias in Hiring},
  year = {2018},
  note = {Reference in text}
}

@article{SocioeconomicAI2019,
  author = {{Socioeconomic Bias in AI Study}},
  title = {Socioeconomic Bias in Algorithmic Systems},
  year = {2019},
  note = {Reference in text}
}

\end{document}